\documentclass{article}

\usepackage[final]{corl_2022} % Uncomment for the camera-ready ``final'' version.
\usepackage{caption} %
\usepackage{hyperref}

\usepackage[textsize=tiny]{todonotes}
\usepackage{amsmath}
\usepackage{amsthm}
\usepackage{xcolor}
\usepackage{graphicx}
\usepackage{cite}
\usepackage{amsmath, amssymb}
\usepackage{mathtools}
\usepackage{soul}
\usepackage{algorithm}
\usepackage{subcaption}
\usepackage{mathrsfs}
\usepackage{todonotes}

\usepackage{amsmath,amsfonts,bm}

\def\eqref#1{equation~\ref{#1}}

\def\1{\bm{1}}

\DeclareMathAlphabet{\mathsfit}{\encodingdefault}{\sfdefault}{m}{sl}
\SetMathAlphabet{\mathsfit}{bold}{\encodingdefault}{\sfdefault}{bx}{n}

\DeclareMathOperator*{\argmax}{arg\,max}

\usepackage[autostyle]{csquotes}
\usepackage{wrapfig}

\theoremstyle{plain}
\newtheorem{theorem}{Theorem}[section]

\theoremstyle{definition}
\newtheorem{definition}[theorem]{Definition}
\newtheorem{assumption}[theorem]{Assumption}
\theoremstyle{remark}

\newcommand\reals{\mathbb{R}}
\newcommand\prob{\mathbb{P}}
\newcommand\optionO{\mathcal{O}}
\newcommand\optionI{I}
\newcommand\optionT{T}
\newcommand\init{\mathcal{I}}

\newcommand\env{\mathcal{E}}

\newcommand\ops{\textit{option templates }}

\definecolor{ed}{RGB}{225,0,0}
\definecolor{sd}{RGB}{0,0,225}
\definecolor{ks}{RGB}{225,100,0}

\newcommand{\bitem}{\begin{itemize}}
\newcommand{\eitem}{\end{itemize}}
\newcommand{\benum}{\begin{enumerate}}
\newcommand{\eenum}{\end{enumerate}}
\newcommand{\beq}{\begin{equation}}
\newcommand{\eeq}{\end{equation}}
\newcommand{\beqs}{\begin{equation*}}
\newcommand{\eeqs}{\end{equation*}}

\usepackage{amsfonts} %

\usepackage{algorithmic} %

\title{Exploring with Sticky Mittens:\\Reinforcement Learning with Expert Interventions via Option Templates}

\author{%
    Souradeep Dutta$^1$, Kaustubh Sridhar$^1$, Osbert Bastani$^1$, Edgar Dobriban$^1$, James Weimer$^2$, \\$\;$\\ \textbf{Insup Lee$^1$, Julia Parish-Morris$^3$}\\$\;$\\
    $^1$University of Pennsylvania, $^2$Vanderbilt University, $^3$Children's Hospital of Philadelphia
}

\begin{document}
\maketitle

\begin{abstract}
% Environments with sparse rewards and long horizons pose a significant challenge for current reinforcement learning algorithms.
Long horizon robot learning tasks with sparse rewards pose a significant challenge for current reinforcement learning algorithms.
A key feature enabling humans to learn challenging control tasks is that they often receive expert intervention that enables them to understand the high-level structure of the task before mastering low-level control actions.
We propose a framework for leveraging expert intervention to solve long-horizon reinforcement learning tasks. We consider \emph{option templates}, which are specifications encoding a potential option that can be trained using reinforcement learning. We formulate expert intervention as allowing the agent to execute option templates  before learning an implementation. This enables them to use an option, before committing costly resources to learning it. 
We evaluate our approach on three challenging reinforcement learning problems, showing that it outperforms state-of-the-art approaches by two orders of magnitude. Videos of trained agents and our code can be found at: \url{https://sites.google.com/view/stickymittens}
\end{abstract}

\keywords{Sample-Efficient Reinforcement Learning, Expert Intervention, Options, Planning with Primitives} 

% \input{introduction_through_child_development}
% Page limit for CoRL - 8
\section{Introduction}
% In reinforcement learning (RL), the agent's goal is to maximize a reward function while exploring its environment. Modern techniques such as deep reinforcement learning can effectively solve this problem in large state spaces in a number of cases~\citep{deep-q-paper}, enabling reinforcement learning to solve difficult tasks such as robot planning and locomotion~\citep{demo_RL} and playing video games from visual inputs~\citep{high_dim_RL}.
Reinforcement learning is an effective tool to solve difficult tasks such as robot planning and locomotion~\citep{demo_RL} but exploration is still a challenge. 
% In order to apply RL effectively to practical applications with a high-dimensional state and action space, exploration is a challenge. Generally, complex sequences of actions are required to achieve any nonzero reward. Hence, random exploration will take extremely long to find a nonzero reward signal. 
\emph{Options} are an RL tool to circumvent this problem~\citep{options_paper}. Designed to achieve intermediate subgoals. For instance, in robot grasping tasks, an option might enable the robot to grasp a block, which is a subgoal needed to build a tower out of blocks. The goal is to learn a policy mapping each state to an option, instead of a concrete action to take.  
% Options naturally guide exploration, making it more likely that the agent encounters nonzero rewards.

When learning to perform complex visual-motor skills, humans often rely on expert interventions to help them escape these challenging reward plateaus. For instance, the \textit{sticky-mittens} experiment~\citep{sticky_mittens_exp} considers infants who have not yet learned to grasp objects. They give a subset of these infants mittens covered with Velcro hooks and allow them to play with toys fitted with Velcro loops, making it significantly easier for them to grasp these toys. Even if the Velcro is taken away, these babies learn how to grasp objects significantly faster than infants not exposed to this experience. In other words, enabling infants to explore unreachable parts of the state space helps guide them towards skills that are worth learning. 
%This extends beyond fine motor skills and is a well known phenomenon in developmental psychology.
This is a well known phenomenon in developmental psychology, which extends beyond fine motor skills.

In this paper, we design an RL algorithm based on the idea from the sticky-mittens experiment. The agent has access to an alternative Markov Decision Process (MDP) where the agent can leap multiple states without first learning a policy to do so in the original MDP. We term such a jumping mechanism an \emph{option template}. Option templates are described using a set of initial states and a set of final states.
% In this paper, we design an RL algorithm based on the idea from the sticky-mittens experiment of providing training time expert help. In particular, our algorithm provides the agent with an alternative Markov Decision Process (MDP) where the agent can leap multiple states without first learning a policy to do so in the original MDP. We term such a jumping mechanism an \emph{option template}. Option templates are described using a set of initial states and a set of final states.  The agent uses reinforcement learning to train an option to implement each option template. Using an option template moves the agent to a state in the desired set of final states. After this, the agent can continue exploration. As a consequence, the agent acquires the ability to use an option before deciding to spend effort to learn it. Option templates enable the agent to predict the effects of options so as to be able to plan over long and difficult tasks. % should probably stay away from the word teleport
The idea of providing external help in the learning phase is referred to as \textit{primitives} or \textit{skills} in literature. It offers a practical way to speed-up learning in a realistic setting. For instance, using parameterized action spaces for RoboCup in \citep{ICLR16-hausknecht}, stitching independent behaviors in \citep{ut_austin_robocup, barreto2019option} and providing action primitives in \citep{action_primitives, icra22_primitves, chitnis2020efficient} to mention a few. 

% Generally, in the more typical framework of training with options, the agent is faced with the challenge of learning options over environmental actions first. Training these low-level policies 
In the more typical RL framework the agent first uses reinforcement learning to train an option to implement the specification of each option template. The issue with this strategy is that, in RL environments with large state spaces the options learnt need to be of a fairly generic nature, in order to be useful. Which is hard without a knowledge of the state distribution, where the options will be invoked. Using option templates we decouple the implementation of an option from its utility.
% Generally, in the more typical framework of training with options, the user is burdened with providing the agent with an implementation of each option. Thus, the agent first uses reinforcement learning to train an option to implement the specification of each option template, after which it can again use reinforcement learning to train a policy over these options. The issue with this strategy is that, in RL environments with large state spaces the options learnt need to be of a fairly generic nature, in order to be useful. Without a knowledge of the state distribution, where the options will be invoked by a higher level policy. Through the use of \emph{option templates} we decouple the implementation of an option from its utility, by providing the agent with a deterministic transition to the outcome of running the option.

% Here, we explore robotic manipulation such as the one described in the Mujoco environment FetchAndStack, where the aim is to learn stacking blocks in the right order, using actions which control joint torques, and gripper force. Another environment we explore is that of GFootball, in a multi-player control setting. The actions available are that of basic kicking force, and motion primitives for each player on field. The aim is to score goals, and win the current episode. Finally we explore an environment -Craft, which is a Minecraft like planning environment to showcase the generality of our approach. The external helping action is mainly for exploration purposes. The agent perfects the policy to perform a task after it has discovered its utility.

In more detail, our algorithm performs an alternating search over policies: at each iteration, it first optimizes the high-level policy over the current option templates, and then learns options to implement the option templates. 
% It continues this iterative procedure until it has learned a policy that uses the actions in the original action space. 
% Once the high-level policy learns from which states to call an option template, it provides an implicit guide to the following level. 
In our experiments, we demonstrate that by leveraging option templates, our algorithm can achieve orders of magnitude reduction in sample complexity compared to the typical strategy of learning the options upfront and then planning with these learned options.

\textbf{Contributions: } Our contributions are: (1) an RL algorithm that leverages option-templates, and (2) an empirical comparison with state-of-the-art RL techniques on three challenging environments, demonstrating order-of-magnitudes reduction in sample complexity.
\vspace{-2mm}
\section{Background}
\vspace{-2mm}
We consider the RL setting where an agent interacts with an environment modeled as an MDP \citep{sutton2018reinforcement}.
\begin{definition}[\textbf{MDP}]
\label{def:mdp}
A \emph{Markov Decision Process (MDP)} is a tuple $\env=(\mathcal{S}, \mathcal{A}, \mathcal{P}, \mathcal{R}, \gamma, \init )$, where  $\mathcal{S} \subseteq \reals^n$ is the set of states, $\mathcal{A} \subseteq \reals^m$ is the set of actions, $\mathcal{P}(s'|s,a)$ is the probability of transitioning from state $s$ to $s'$ upon taking action $a$, $\mathcal{R}(s,a)$ is the reward accrued in state $s$ upon taking action $a$, $\gamma \in [0,1)$ is the discount factor, and $\init$ is the initial state distribution.
\end{definition}

% The dynamics evolve over discrete time steps, where the agent transitions from state $s_t$ to $s_{t+1}\sim\mathcal{P}(\cdot|s_t,a_t)$ by taking an action $a_t$, receiving a reward $r_t=\mathcal{R}(s_t,a_t)$.  
% The \emph{return} at time $t$ from state $s_t$ is the sum of discounted future rewards $\mathbf{R}(s_t) = \sum_{j=t}^{\infty} \gamma^{j-t}r(s_j,a_j)$. A deterministic policy $\pi$ is a mapping $\pi : \mathcal{S} \rightarrow \mathcal{A}$. 

\begin{wrapfigure}{r}{0.3\textwidth}
\vspace{-3mm}
\begin{center}
\includegraphics[width=4cm, height=1.8cm]{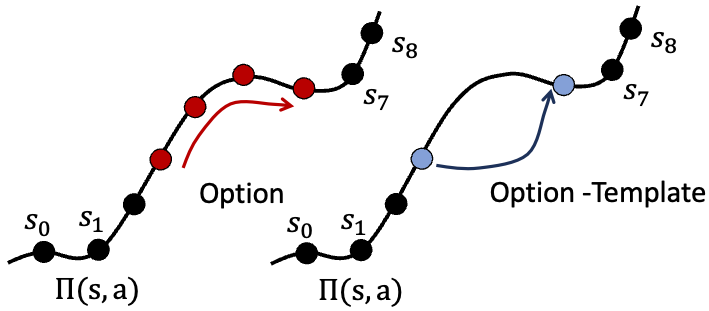}
\end{center}
\caption{Option vs Option Template.}
\label{fig:uber_actions}
\vspace{-6mm}
\end{wrapfigure}
The value of a state for policy $\pi$ is the expected return $\mathbf{R}_\pi(s_0)$ starting from state $s_0$, while executing the policy $\pi$ at every step, $a_t = \pi(s_t)$. The optimal policy $\pi^*$ maximizes the reward starting from the initial state distribution--i.e., $\pi^*$ $=$ $\argmax  V^{\init}(\pi)$,
where $V^{\init}(\pi)$ $=$ $\mathbb{E}_{s_0 \in \init} [\mathbf{R}_\pi(s_0)]$. Semi-MDPs (SMDPs) extend this framework by allowing temporally extended actions---i.e., policies executed over multiple steps---called \emph{options}~\citep{options_paper}.
\begin{definition}[\textbf{Option}]
\label{defn:option}
An \emph{option} $o$ is a tuple $o :=(\optionI, \optionT, \pi)$, 
where $\optionI \subseteq S$ are initial states, $\optionT: \mathcal{S} \rightarrow [0,1]$  the termination condition, and $\pi: \mathcal{S} \rightarrow \mathcal{A}$ is a policy.
\end{definition}

For now, assume that we have a set $\optionO$ of options available for an MDP. 
%  Intuitively, options constrain the search space over policies, thereby guiding exploration performed by the agent towards more promising action sequences. In particular, we now consider policies $\Pi:\mathcal{S}\to\optionO$ that map a state $s\in\optionI$ (i.e., $s$ is a valid initial state for $o$) to an option $o=\Pi(s)$. Then, the agent uses policy $\pi$ until the termination condition holds. More precisely, after taking action $a=\pi(s)$ and transitioning to state $s'\sim\mathcal{P}(\cdot|s,a)$, it stops using $\pi$ with probability $\optionT(s_{t+1})$ and chooses another option $o'=\Pi(s')$; otherwise, it continues using $\pi$.
 Intuitively, options constrain the search space over policies, thereby guiding exploration performed by the agent towards more promising action sequences.  We now consider policies $\Pi:\mathcal{S}\to\optionO$ that map a state $s\in\optionI$ (i.e., $s$ is a valid initial state for $o$) to an option $o=\Pi(s)$. The agent uses option policy $\pi$ until the termination condition holds. More precisely, after taking action $a=\pi(s)$ and transitioning to state $s'\sim\mathcal{P}(\cdot|s,a)$, it stops using $\pi$ with probability $\optionT(s_{t+1})$ and chooses another option $o'=\Pi(s')$; otherwise, it continues using $\pi$.

% For Markovian options, the termination condition $\optionT$ or the option sub-policy $\pi$ depends only on the current state only. In practice, it is useful to have the ability to \textit{time out}, i.e., to terminate an option even if it does not satisfy some desirable end goal. Suppose that an option $o$ is invoked at time $t$ in state $s_t$, and assume that under the control of this option, the system follows a $k$ step trajectory $ \mathsf{Tr}^{o}(s_t, k)=(s_{t}, a_{t}, r_{t}, s_{t+1}, \dots , s_{t+k})$.
% It is possible that for $t \leq i < t + k$, the action $a_i$ depends on the entire history $\mathsf{Tr}^{o}(s_t, i)$ from $t$ until $t + i$.  This use of memory renders the behavior non-Markovian. Denoting the set of all such state-action trajectories as $H$, we can allow option policies $\pi: H \rightarrow \mathcal{A}$, and termination conditions $\optionT: H \rightarrow [0,1]$ to be semi-Markovian.
Suppose that an option $o$ is invoked at time $t$ in state $s_t$, and the system follows a $k$ step trajectory $ \mathsf{Tr}^{o}(s_t, k)=(s_{t}, a_{t}, r_{t}, s_{t+1}, \dots , s_{t+k})$. We consider policies such that for $t \leq i < t + k$, the action $a_i$ depends on the entire history $\mathsf{Tr}^{o}(s_t, i)$ from $t$ until $t + i$. This renders the behavior non-Markovian. Denoting the set of all such state-action trajectories as $H$, we can allow option policies $\pi: H \rightarrow \mathcal{A}$, and termination conditions $\optionT: H \rightarrow [0,1]$ to be semi-Markovian.
Now, we recall the definition of Semi-MDPs as an extension of MDPs:

\begin{definition}[\textbf{Semi-MDP},  \citet{options_paper}]
A semi-MDP is a tuple $(\mathcal{S}, \mathcal{O}, \mathcal{P}, \mathcal{R}, \gamma, \init )$, where $\mathcal{S}$ is a set of states, $\mathcal{O}$ is a set of semi-Markovian options, $\mathcal{P}$ is the transition probability between states, $\mathcal{R}$ is the reward function, $\gamma \in [0,1)$ is the discount factor, and $\init$ is the initial state distribution.
\end{definition}

The state-prediction part of the model for taking an option $o$, in state $s \in \optionI $ and transitioning to $s'$, is given by \citet{options_paper}: $ P(s'| s, o) = \sum_{k=1}^{\infty} \gamma^k p(s',s,k)$, 
where $p(s',s,k)$ is the probability that the option terminates in $s'$ after $k$ steps from the time the option is invoked. 
% Let $\mathscr{E}(o,s,t)$ be the event of taking an option $o$ while in state $s$ at time $t$. Then the time-discounted reward is $ \mathbf{R}(s, o) = \mathbb{E}\{r_{t} + \gamma r_{t+1} + \dots + \gamma^k r_{t+k} | \mathscr{E}(o,s,t) \} $, where $t+k$ is the random time at which the option $o$ terminates. 
The option-value form of standard value iteration is, for $s \in \mathcal{S}$ and $o \in \mathcal{O}$, \citep[][eq. 12]{options_paper},
\citep{Barto03recentadvances}
% \begin{equation}
%     \label{eq:option-value}
$    Q^{*}_{\optionO}(s, o) = R(s, o) + \sum_{s' \in S}P(s'|s, o) \max_{o' \in \optionO} Q^*_{\optionO}(s', o'),$
% \end{equation}
where $s'$ is the state the system reaches after executing option $o$ starting from state $s$. SMDP value learning updates the $Q$-values at the end of each option termination as  \citep[][p. 195]{options_paper}:
% \begin{equation}
    % \label{eq:value-update}
$    Q^{i+1}(s, o) \leftarrow Q^i(s, o) + \alpha[r_c + \gamma^k \max_{o' \in \optionO }Q^i(s', o') - Q^i(s, o)],$
% \end{equation}
for similar definitions of $k$, $s$, $s'$, with $r_c$ being the cumulative discounted reward over the time horizon, and $\alpha \in (0,1)$ the learning rate. 
Choosing a function approximator parameterized by $\theta$ to represent $Q(s, o; \theta)$, with a loss given by :  
$L(\theta_i) = [r + \gamma^k \max_{o' \in O}Q^i(s', o';\theta_i) - Q^i(s, o; \theta_i)]^2$,
allows one to use standard Q-learning \citep{deep-q-paper} methods to train with options.
\vspace{-2mm}

% \input{uber_actions}
% \vspace{-2mm}
\section{Learning with Option Templates}

\subsection{Option Templates}
\vspace{-1mm}
% A key challenge with options is that it is often difficulty to supply the option policy $\pi$; while this policy can be learned, it can be hard for the agent to judge the starting state distribution.
Motivated by the sticky-mittens experiment, we consider an environment where in certain states, the agent can call a ``help switch'', called an \emph{option template}, that immediately transitions it from a state $s_t$ to a different state $s_{t+k}$; this transition captures the desired result of executing an (unimplemented) option. We denote a trajectory of this MDP starting from state $s_t$ under policy $\pi$ by $\mathsf{Tr}^{\pi}(s_t)$. 

% \vspace{-1mm}

\begin{definition}[\textbf{Option template}]
\label{defn:option_template}
Option template $a_o$ is a tuple $(\optionI, \optionT, P) $, where $\optionI \subseteq \mathcal{S}$ is the initial states, $\optionT: \mathcal{S} \rightarrow [0,1]$ is the termination condition, and $P$ is a distribution over states $s$ such that $\optionT(s)=1$.
\end{definition}

% \vspace{-1mm}

An option template similar to an option, shifts control to the expert until the termination condition is reached; this termination condition $\optionT$ is intended to capture the satisfaction of some sub-goal along with a timeout mechanism. Assume the termination condition at time $t+k$ of an option template taken in state $s_t$ is of the form $\optionT(s_t, k) = \mathbf{F}_{o}(s_{t+k}) \vee (k > k^*)$, where the \emph{function} $\mathbf{F}_{o}: \mathcal{S} \rightarrow \{0,1\}$ captures visitation of some key states, and $k^*$ is an upper limit on the number of steps.

% Furthermore, the distribution $P$ enables us to ``teleport'' to a state that satisfies the termination condition---i.e., if an agent invokes $a_o$ from a state $s\in I$, then we sample a terminal state $s'\sim T$ and transition the MDP to $s'$.

We denote the set (or subset) of option templates by $\mathcal{A_\optionO}$. By the end of training, the agent must learn a policy to implement each option template that it uses (i.e., the final policy cannot depend on option templates). To ensure feasiblity, we assume there is an a priori unknown policy $\pi_{o}$ such that $\mathsf{Tr}^{\pi_{o}}(s_t)$ satisfies $\optionT$ (denoted as $\mathsf{Tr}^{\pi_{o}}(s_t) \models \optionT$) with probability at least $1-\delta$, for some hyperparameter $\delta \in [0,1)$. Formally, the probability of success of a given policy $\pi_o$ can be expressed as $\prob((\pi_o, \optionI, \env)\models \optionT) \ge (1-\delta)$---i.e., the trajectory starting from a state in $\optionI$ under the control of $\pi_o$ in environment $\env$ satisfies $\optionT$ with probability at least $1-\delta$. This success probability can be estimated by empirical rollouts of $ \pi_{o}$.

% Next, there are different degrees of help that option templates can offer; option templates that traverse longer sequences of transitions can help the agent learn a high-level strategy more quickly since they need to make fewer decisions, but learning their implementations can be more costly. Thus, option templates can be organized in a hierarchical fashion, where longer-horizon option templates can themselves be implemented in terms of several shorter-horizon option templates.

We organize option-templates in a hierarchical fashion depending on degrees of help. Option templates that traverse longer sequences of transitions allow faster learning due to fewer decisions. Thus, we consider a sequence of environments $\env^0, \env^1, \dots , \env^{n-1} = \env$, equipped with option templates of varying capabilities organized in a hierarchical fashion. 

\begin{definition} [\textbf{Environment Level $\env^l$}]
An environment $\env^l$ at learning level $l$ is the original environment $\env$ with the primitive actions $\mathcal{A}$ replaced by option templates $\mathcal{A}_{\optionO}^l$.
\end{definition}
The goal of the levels is to gradually increase difficulty of the learning task while guiding the agent at each level in a way that is similar to curriculum learning \citep{curriculum_learning}. In particular, learning longer timescale option templates typically precedes learning shorter ones.

% Extending our above feasibility assumption, we additionally assume that for neighboring environment levels, option templates in one level can be faithfully captured by the policies in the next level.
\begin{assumption}[\textbf{Realizability}]
For any option template
$a_o \in \mathcal{A}_{\optionO}^l$
at level $l$, given by $(\optionI, \optionT, P)$, there is a policy $\pi^{a_o}: \mathcal{S} \rightarrow \mathcal{A}_{\optionO}^{l+1}$ at the following level $l+1$ such that for any $s_0 \in \optionI$,  $\mathsf{Tr}^{o}(s_0) \models \optionT$ 
with probability at least $1-\delta$.
\end{assumption}
\vspace{-2mm}
That is, any option template in the current level is realizable by a policy in the following level.
\vspace{-1mm}
\subsection{Learning with Option Templates}
\vspace{-1mm}
Our algorithm for RL with option templates is presented in Algorithm \ref{alg:uber_learning}.

\begin{minipage}[t]{0.46\textwidth}
\vspace{-5mm}
\begin{algorithm}[H]
\caption{Learning with Option Templates}
\label{alg:uber_learning}
\textbf{Input: } Environments  $[ \mathcal{E}^0, \mathcal{E}^1, \dots , \mathcal{E}^{n-1} ]$

\textbf{Output: } A set of options $\mathcal{D}=\{o^q\mid q\in Q\}$ that provides an implementation of each option template $a_o^q$, where $q\in Q$ indexes the option templates.

\begin{algorithmic}[1]
\STATE Initialize $\mathcal{D} = \{ \}$
\FOR{level $l \in \{ 1, \dots, n \} $ }
\STATE Let $\mathcal{A}_o^{l-1}$ be the set of option templates on level $l-1$
\STATE Build reward functions $\{ \mathfrak{R}_q \mid a_o^q\in\mathcal{A}_o^{l-1}\} $ 
\FOR{$ a_o^q \in \mathcal{A}_o^{l-1}$ }
\STATE $\pi_o^q=$LearnOptionPolicy$(\mathcal{E}^{l}, \mathfrak{R}_q, a_o^q, \mathcal{D})$ 
\STATE $o^q := $  $(a_o^q, \pi_o^q)$ 
\STATE $\mathcal{D} = \mathcal{D} \cup \{o^q\}$ 
\ENDFOR
\ENDFOR
\OUTPUT $\mathcal{D}$ 
\end{algorithmic}
\end{algorithm}
\begin{figure}[H]
\includegraphics[width=\linewidth]{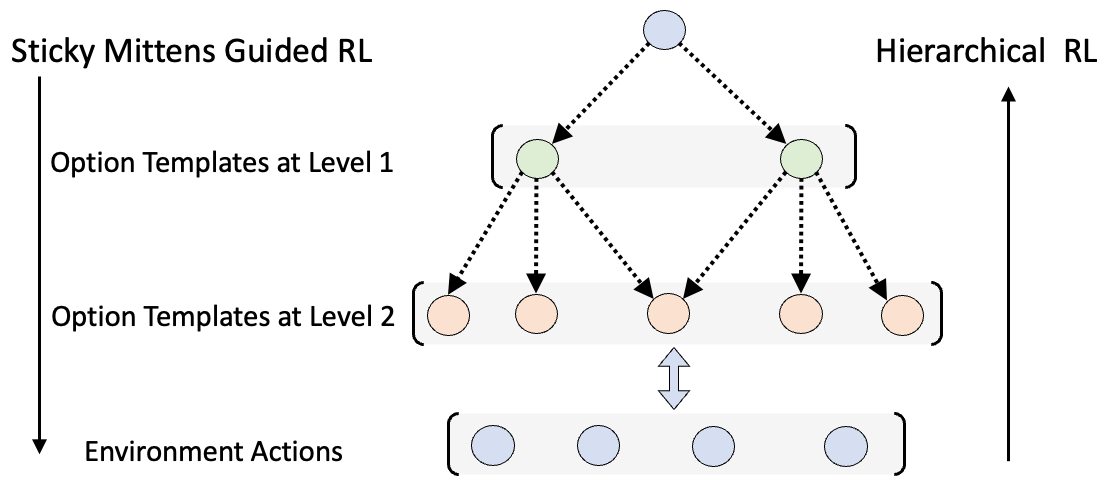}
\caption{Visualization of learning with option templates.}
\label{fig:sticky-mittens-comp}
\end{figure}
\end{minipage} 
\hfill
\begin{minipage}[t]{0.46\textwidth}
\vspace{-5mm}
\begin{algorithm}[H]
\caption{Learn Option Policy}
\label{alg:guided_learning}
\textbf{Input: } Environment  $ \mathcal{E}$, Reward Function $\mathfrak{R}$, Option template $a_o=(I,T,P)$, Set $\mathcal{D}$ of options \\
\textbf{Parameter: } Threshold $\delta$, Max Episodes $L$ \\
\textbf{Output: } Policy $\pi_o$ as an implementation of $a_o$
\begin{algorithmic}[1]
\STATE Initialize policy $\pi_o$
\FOR { episode $= 1, \dots, L$ }
\STATE \textit{ExecutionStack} $= [ ]$
\STATE $o$ $\leftarrow$ Top-level option from $\mathcal{D}$ (or $a_o$ if $\mathcal{D}=\varnothing$)
\STATE Push $o$ onto \textit{ExecutionStack}
\STATE Sample initial state $s\sim I$
\WHILE {not done}
\STATE $o\leftarrow$ Pull option from \textit{ExecutionStack} \\
\STATE $o'\leftarrow\pi_{o}(s)$
\IF{IsOption$(o', a_o)$}
\STATE $(\pi_o, s)\leftarrow$ StepAndTrain$(\mathcal{E},\mathfrak{R},\pi_o,s)$
\ELSIF {$\neg$HasImplementation$(o',\mathcal{D})$ } 
\STATE $s\gets$ Teleport$(o',s)$
\ELSE
\STATE Push $o'$ to \textit{ExecutionStack} 
\ENDIF
\ENDWHILE
\STATE Exit loop if AvgReward $\ge1-\delta$
\ENDFOR
\OUTPUT $\pi_o$
\end{algorithmic}
\end{algorithm}
\end{minipage}

% \begin{wrapfigure}{r}{0.35\textwidth}
% \begin{center}
% \includegraphics[width=5cm, height=2.5cm]{Figures/sticky_mittens_learning.png}
% \end{center}
% \caption{Visualization of learning with option templates.}
% \label{fig:sticky-mittens-comp}
% \vspace{-5mm}
% \end{wrapfigure}

\vspace{2mm}

Intuitively, at each iteration, it ``flattens'' the option templates used on the previous level $l-1$ based on the ones available at the current level $l$. At each level $l$, we first design reward functions $\mathfrak{R}_q$ for each option template $a_o^q\in\mathcal{A}_o^{l-1}$ (Line 4); this reward function is later used to learn a policy $\pi_o^q$ implementing $a_o^q$. For the base case $l=1$, we take the reward function $\mathfrak{R}_q$ to be the reward function for the original environment $\mathcal{E}$; thus, learning the policy for the option-template at this level amounts to accomplishing the goals of $\mathcal{E}$ using the (unique) option template $a_o^q\in\mathcal{A}_o^0$. For subsequent iterations $l>1$, $\mathfrak{R}_q$ encodes the goal of achieving the termination condition of $a_o^q$, based on its termination condition (Line 4). 

Given $\mathfrak{R}_q$, our algorithm learns a policy that maximizes $\mathfrak{R}_q$ using the options available at the current level $l$ (Line 6). For $l=1$, we assume there is a single option template $\mathcal{A}_o^0=\{a_o^q\}$; the corresponding policy $\pi_o^q$ aims to achieve the goal in the original environment $\mathcal{E}$, so we refer to the resulting option $o^q=(a_o^q,\pi_o^q)$ as the \emph{top-level option}. For $l>1$, $\pi_o^q$ implements an option template $a_o^q\in\mathcal{A}_o^{l-1}$ at level $l-1$ using the option templates available at the current level $l$. Importantly, this process leverages policies learned so far to generate initial states from which to learn $\pi_o^q$. Once we have learned $\pi_o^q$, we add the resulting option $o^q=(a_o^q,\pi_o^q)$ to $\mathcal{D}$.
% \begin{algorithm}[t]
% \caption{Learn Option Policy}
% \label{alg:guided_learning}
% \textbf{Input: } Environment  $ \mathcal{E}$, Reward Function $\mathfrak{R}$, Option template $a_o=(I,T,P)$, Set $\mathcal{D}$ of options \\
% \textbf{Parameter: } Threshold $\delta$, Max Episodes $L$ \\
% \textbf{Output: } Policy $\pi_o$ as an implementation of $a_o$

% \begin{algorithmic}[1]
% \STATE Initialize policy $\pi_o$
% \FOR { episode $= 1, \dots, L$ }
% \STATE \textit{ExecutionStack} $= [ ]$
% \STATE $o$ $\leftarrow$ Top-level option from $\mathcal{D}$ (or $a_o$ if $\mathcal{D}=\varnothing$)
% \STATE Push $o$ onto \textit{ExecutionStack}
% \STATE Sample initial state $s\sim I$
% \WHILE {not done}
% \STATE $o\leftarrow$ Pull option from \textit{ExecutionStack} \\
% \STATE $o'\leftarrow\pi_{o}(s)$
% \IF{IsOption$(o', a_o)$}
% \STATE $(\pi_o, s)\leftarrow$ StepAndTrain$(\mathcal{E},\mathfrak{R},\pi_o,s)$
% \ELSIF {$\neg$HasImplementation$(o',\mathcal{D})$ } 
% \STATE $s\gets$ Teleport$(o',s)$
% \ELSE
% \STATE Push $o'$ to \textit{ExecutionStack} 
% \ENDIF
% \ENDWHILE
% \STATE Exit loop if AvgReward $\ge1-\delta$
% \ENDFOR
% \OUTPUT $\pi_o$
% \end{algorithmic}
% \end{algorithm}
% \textbf{Reward function construction.}
Recall that the termination condition for an option starting at state $s_0$ is $\optionT(s_0, k^*) = \mathbf{F}_{o}(s_{k}) \vee (k > k^*)$. The goal of training $\pi_o^q$ is to satisfy this condition with high probability.
Hence, when training policy $\pi_o^q$ for option $a_o^q$, we choose the reward function $\mathfrak{R}_q$ 
to be $\mathbf{F}_{o}$, and additionally restrict the length of each episode to $k^*$ time-steps.

\textbf{Learning option policies.} Next, we describe LearnOptionPolicy, which Algorithm \ref{alg:guided_learning} uses to learn a policy for option template $a_o^q$. For simplicity, we denote the current environment by $\mathcal{E}$, the current reward function by $\mathfrak{R}$,
the option template by $a_o$,
and the target policy by $\pi_o$. The goal of $\pi_o$ is to achieve the termination condition $T$ for $a_o=(I,T,P)$ from initial states $s\in I$. 

One challenge is sampling an initial state $s\in I$ from which to train $\pi_o$. To do so, this subroutine leverages access to the previously learned options $\mathcal{D}$; it uses options in $\mathcal{D}$ until it arrives at a state $s\in I$ where $a_o$ is called. In more detail, it samples $s$ by executing the top-level option in $\mathcal{D}$ until it reaches a state where $a_o$ is called. In general, executing an option $o$ either relies on executing its policy $\pi_o$ in $\mathcal{D}$ if it exists, or by executing its ``teleport'' functionality. Our algorithm keeps track of the execution of options for which $\pi_o$ exists using a stack, which is initialized with the top-level option (Line 3-5). The structure is visualized in Figure~\ref{fig:sticky-mittens-comp}. Then, while executing the current option $o$, it obtains the next option $o'$ to execute, which is processed in one of three ways: \vspace{-2mm}
\begin{itemize}
\item If $o'$ is the option for $a_o$ (i.e., it has a matching initial state, termination condition, and a current policy $\tilde \pi_o$, checked by IsOption on Line 10), then it takes steps to train $\pi_o$ (Line 11); as described below. 
\item If $o'$ does not have an implementation (checked by HasImplementation on Line 12), then it teleports---i.e., we sample $s\sim P$ (Line 13).
\item Otherwise, $o'$ has an implementation, so it pushes $o'$ onto the stack (Line 15). 
\end{itemize}
\vspace{-1mm}
% In general, $o'$ will have an implementation if it is on a level smaller than $l-1$ (since these were learned on previous iterations of Algorithm~\ref{alg:uber_learning}), and will not have an implementation if it is on level $l$; if it is on level $l-1$, then it may or may not have an implementation depending on the order in which policies are learned. This structure is visualized in Figure~\ref{fig:sticky-mittens-comp}.
\begin{figure*}[t!]
    \begin{subfigure}{0.29\textwidth}
        \centering
        \includegraphics[width=\linewidth]{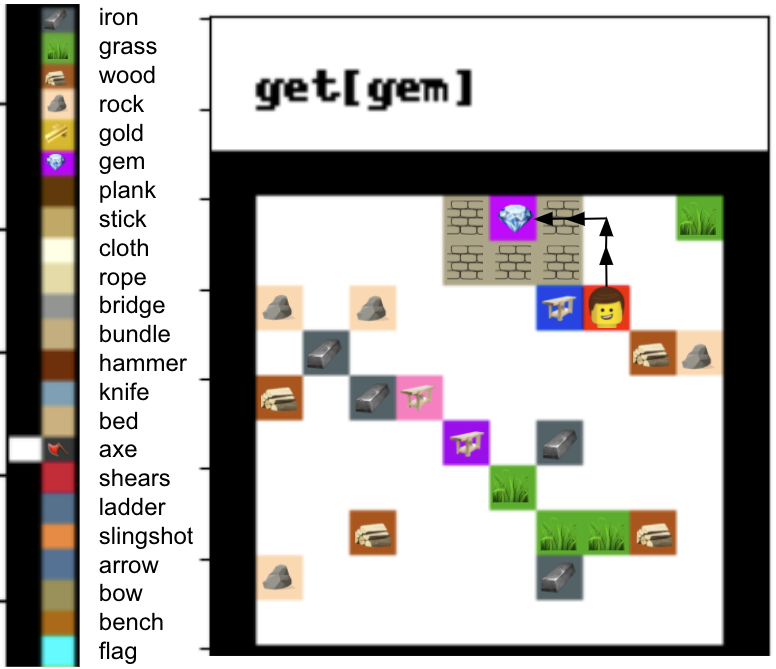}
        \caption{Visualization of craft environment for the \textit{get gem} task: The agent (red square) uses the axe in its inventory to break the stone and retrieve the gem.}
        \label{fig:craft}
    \end{subfigure} \hfill 
    \begin{subfigure}{0.29\textwidth}
        \centering
        \includegraphics[width=\linewidth]{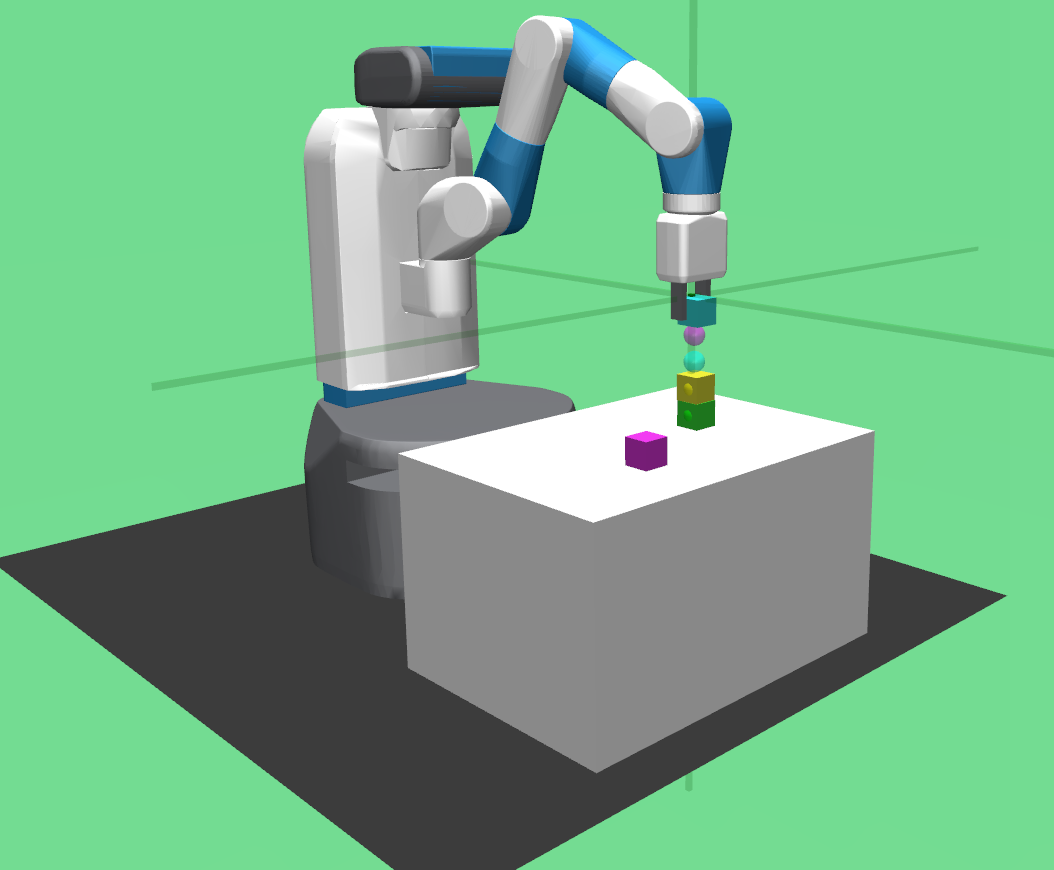}
        \caption{Visualization of fetch and stack environment: Each block has to be placed at the location represented by the color-coded sphere.}
        \label{fig:fetch}
    \end{subfigure} \hfill 
    \begin{subfigure}{0.375\textwidth}
        \centering
        \includegraphics[width=\linewidth]{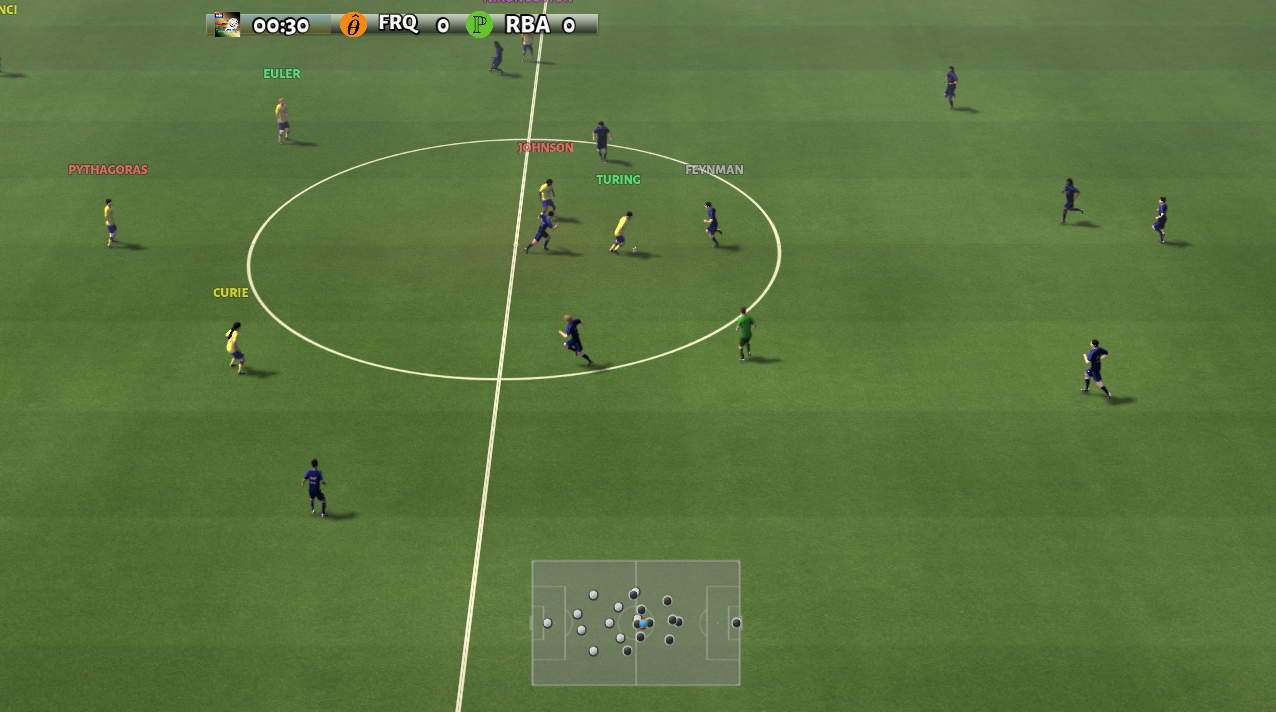}
        \caption{Visualization of the Google football environment: The agents learn to use 11 players of the left team and score goals.}
        \label{fig:gfootball}
    \end{subfigure}
    \caption{Craft, Fetch and GFootball environments.}
    \label{fig:env_pictures}
\end{figure*}
\vspace{-1mm}
Finally, once we are at a state $s$ where $a_o$ is called, the subroutine $StepAndTrain$ takes control and trains $\pi_o$ using a standard reinforcement learning algorithm---i.e., by collecting rewards from $s$ and taking a policy gradient step once it reaches a terminal state or hits the timeout of $k^*$ steps. Then, the current episode continues from the state reached.Once the current episode reaches a final state, then the algorithm continues to the next episode. This process continues until the average reward of $\pi_o$ exceeds a threshold $1-\delta$, or until a maximum number of episodes $L$ is reached.

\section{Experiments}

\label{sec:exp}
\subsection{Experiments on Planning tasks in the Craft Environment}
\label{sec:craft_exp}
\vspace{-2mm}
\textbf{Description of environment: }
The \emph{craft} environment \citep{policy_sketches_paper, craft_env_github_visualization_repo} is a 2D world based on the Minecraft game, where an agent has to complete various hierarchical tasks with sparse rewards. The environment is represented by a $12\times 12$ grid with cells containing raw resources (\textit{e.g., WOOD, IRON}), crafting areas (\textit{e.g., WORKBENCH}), obstacles (\textit{e.g., STONE, WATER}) and valuable items (GOLD or GEM). 
There are four move actions (\textit{up, down, left, right}) and a special \textit{USE} action. To grab an item, the agent moves to a neighboring cell and applies the \emph{USE} action.  Table \ref{table:craft_tasks_and_uber_actions} shows the hierarchical arrangement of the \textit{get gem} task; the details for \emph{get gold} are in the Appendix.
% To grab an item, the agent has to move to a cell next to it, orient itself in its direction, and apply the \emph{USE} action to take it into its inventory. With the necessary items in the inventory, the agent can move to a crafting area (\textit{e.g.}, \textit{WORKBENCH}) where it can use items from the inventory to make a tool (\textit{e.g.}, \textit{AXE}). The tool can then be used to retrieve a valuable item hidden behind a previously inaccessible part of the environment (\textit{e.g.}, using an \emph{AXE} to break \textit{STONE} to retrieve a \emph{GEM}). Table \ref{table:craft_tasks_and_uber_actions} shows all the tasks in the hierarchy of the \textit{get gem} task; the details for \emph{get gold} are in the Appendix.

\textbf{Option template at each environment:}
Our option templates (in Table~\ref{table:craft_tasks_and_uber_actions}) capture the hierarchy of task-dependencies in the environment. For example, it is easier to get \textit{GEM} when the agent has access to \textit{AXE} (see Table~\ref{table:craft_tasks_and_uber_actions}). Thus, at the topmost level the agent can ``ask'' for an \textit{AXE}, and via primitive actions, use it to break stone and get \textit{GEM} to realize the importance of an \textit{AXE}.

\begin{table*}[t]
\begin{minipage}{.4\linewidth}
    \centering
    \begin{tabular}{c|c}
    \hline
    \footnotesize{Task}        & \footnotesize{Policy Sketch}                                                 \\ 
    \hline \hline 
    \footnotesize{get wood}    & -                                                             \\
    \footnotesize{get iron}    & -                                                             \\ \hline
    \footnotesize{make stick}  & \footnotesize{get wood $\rightarrow$ use anvil}                              \\ \hline
    \footnotesize{make axe}    & \footnotesize{make stick $\rightarrow$ get iron}                             \\ 
                & \footnotesize{$\rightarrow$ use workbench}                                   \\ \hline 
    \footnotesize{get gem}     & \footnotesize{make axe $\rightarrow$ break stone}                             \\ \hline
    \end{tabular}
\end{minipage}
\begin{minipage}{.55\linewidth}
    \centering
    \begin{tabular}{c|c|l}
    \hline
    \footnotesize{Task}       & \footnotesize{Learning} & \footnotesize{option templates}           \\ 
                & level         &                       \\
    \hline \hline
    \footnotesize{get gem}    & \footnotesize{1}              & \footnotesize{\{give axe, primitive actions\}}           \\ \hline
    \footnotesize{make axe}   & \footnotesize{2}              & \footnotesize{\{give stick, give iron, primitive actions\}} \\ \hline
    \footnotesize{make stick} & \footnotesize{3}              & \footnotesize{\{give wood, primitive actions\}}             \\
    \footnotesize{get iron}   & \footnotesize{3}              & \footnotesize{\{primitive actions only\}}                     \\ \hline
    \footnotesize{get wood}   & \footnotesize{4}              & \footnotesize{\{primitive actions only\}}                    \\ \hline
    \end{tabular}
\end{minipage}
\caption{Get gem hierarchical task: \textbf{[LEFT]} policy sketches
\citep{policy_sketches_paper} \& \textbf{[RIGHT]} option templates for each task in the hierarchy. The order of the rows represent the learning order in the two alternatives.} 

\vspace{-1em}
\label{table:craft_tasks_and_uber_actions}
\end{table*}

\textbf{Implementation: }In each task of each learning level, we use a vanilla actor-critic algorithm ~\citep{main_actor_critic} with a network with one hidden layer of 100 neurons for both the actor and critic. The input to the network is a feature vector consisting of one-hot encodings of the items in each cell in a $5\times5$ grid around the agent along with a one-hot encoding of its inventory. At the topmost level, the agent is given a reward of 1 if and only if \textit{GOLD} or \textit{GEM} are obtained. After training, we obtain a actor network for individual tasks at different learning levels.

\textbf{Comparison with option-value iteration: }We implement option value iteration ~\citep{options_paper} as a comparison with standard hierarchical learning. In option value iteration, we learn options bottom-up, using options learnt at a lower level to accomplish sub-tasks of options at the level above. That is the agent learns to implement lower level sub-tasks first before it learns how to use them. We plot average reward as a function of episodes for each sub-task of \textit{get gem} in Figure \ref{fig:plots_craft}; for \textit{get gold} the Figure is in Appendix \ref{app:craft}. Figure \ref{fig:plots_craft} shows that option templates obtain higher average rewards than option-value iteration at all levels (detailed discussion in Appendix \ref{app:craft}).

% \begin{table}[t!]
% \centering
% \begin{tabular}{c|c|c}
% \footnotesize{Task}         & \multicolumn{2}{c}{\footnotesize{Episodes}}                                                              \\ 
% \cline{2-3} & \footnotesize{Curriculum learning} & \footnotesize{Option templates} \\
%  & \footnotesize{~\citep{policy_sketches_paper}} &  \\ 
% \hline \hline 
% \footnotesize{get gem} & \footnotesize{$>3 \times 10^6$}  & \footnotesize{12826.0 \scriptsize{$\pm$ 2613.0}}\\      
% \footnotesize{make axe} & \footnotesize{$>2.7 \times 10^6$} & \footnotesize{11283.0 \scriptsize{$\pm$ 2255.0}}\\
% \footnotesize{make stick} & \footnotesize{$>1.3 \times 10^6$}  & \footnotesize{5026.0 \scriptsize{$\pm$ 2231.0}}\\ \hline 
% \end{tabular}
% \caption{Comparison of total episodes (and standard deviations over ten random seeds) to train an agent to solve the \textit{get gem} task via option templates and curriculum learning~\citep{policy_sketches_paper}.}
% \label{table:craft_results}
% \end{table}

\begin{wraptable}{r}{70mm}
\centering
\begin{tabular}{c|c|c}
\footnotesize{Task}         & \multicolumn{2}{c}{\footnotesize{Episodes}}                                                              \\ 
\cline{2-3} & \footnotesize{Curriculum learning} & \footnotesize{Option templates} \\
 & \footnotesize{~\citep{policy_sketches_paper}} &  \\ 
\hline \hline 
\footnotesize{get gem} & \footnotesize{$>3 \times 10^6$}  & \footnotesize{12826.0 \scriptsize{$\pm$ 2613.0}}\\      
\footnotesize{make axe} & \footnotesize{$>2.7 \times 10^6$} & \footnotesize{11283.0 \scriptsize{$\pm$ 2255.0}}\\
\footnotesize{make stick} & \footnotesize{$>1.3 \times 10^6$}  & \footnotesize{5026.0 \scriptsize{$\pm$ 2231.0}}\\ \hline 
\end{tabular}
\caption{Comparison of total episodes (and standard deviations over ten random seeds) to train an agent to solve the \textit{get gem} task via option templates and curriculum learning~\citep{policy_sketches_paper}.}
\label{table:craft_results}
\end{wraptable}

\textbf{Comparison with curriculum learning \citep{policy_sketches_paper}:} Additionally, we compare our method with the curriculum learning algorithm employed in \citet{policy_sketches_paper} and observe a $100$ fold decrease in the number of episodes required to train an agent for the \textit{get gem} and \textit{get gold} tasks (see Table \ref{table:craft_results} for \textit{get gem} and Appendix for \textit{get gold}). For the proposed method, we report the total episodes it requires for average reward to stay consistently above $0.8$. Averaged over ten runs for different random seeds. The calculation of the total episodes for each task, includes all the sub-tasks in its hierarchy. For instance, the episodes for \textit{get gem} completion via option templates include episodes required for completing \textit{make axe}, \textit{make stick}, \textit{get wood} and \textit{get iron}. The latter two  tasks are straightforward, and not included in the table. 

\begin{figure*}[t]
\centering
\includegraphics[width=.32\textwidth]{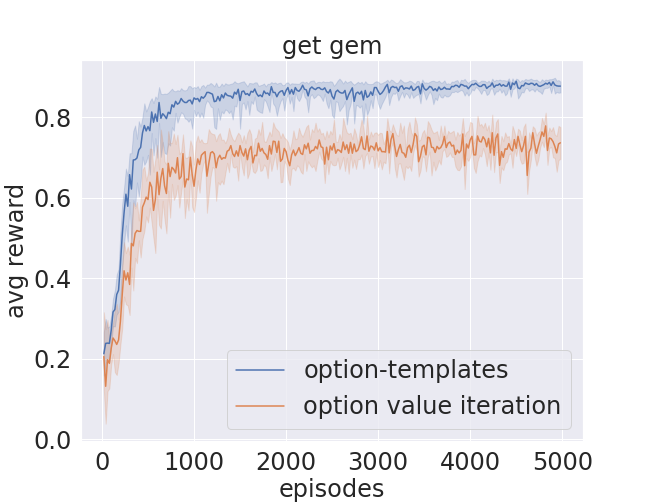}\hfill
\includegraphics[width=.32\textwidth]{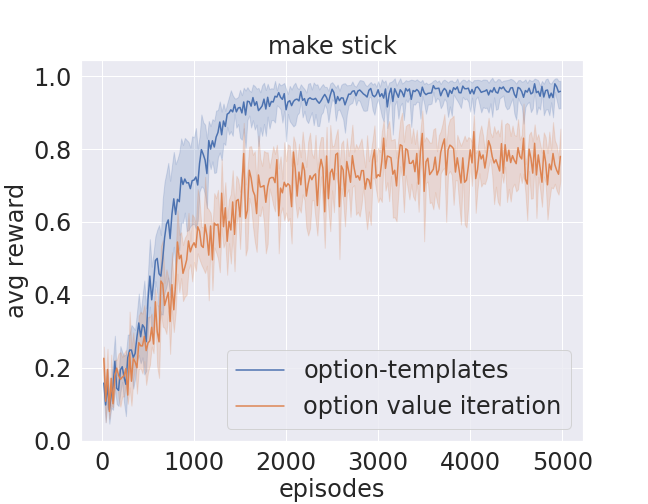}\hfill
\includegraphics[width=.32\textwidth]{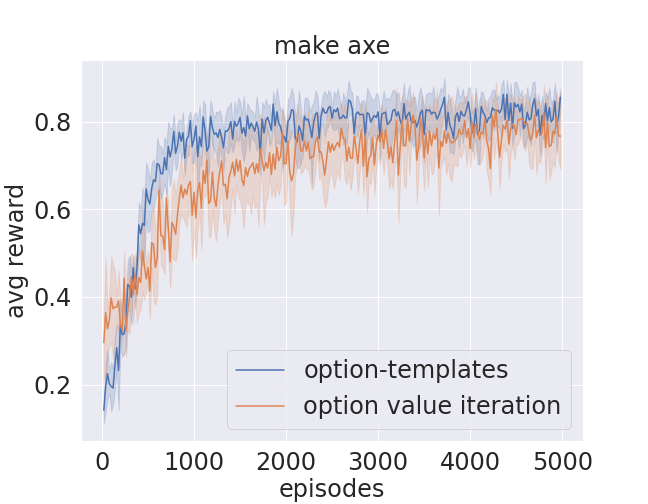}
\caption{Average reward vs episodes for solving each hierarchical sub-task. We compare option templates (ours) and option-value iteration (baseline) for the task \textit{get gem} in craft environment.} 
\vspace{-2.5mm}
\label{fig:plots_craft}
\end{figure*}

\subsection{Experiments on Manipulation Tasks in the Fetch and Stack Environment}
\vspace{-2mm}
\textbf{Description of environment:} This continuous action space environment, introduced in \citet{gym_fetch_stack}, consists of a robotic arm and a platform with blocks placed on it in random positions; see Figure~\ref{fig:fetch}. The goal is for the arm to move, lift, and release the colored blocks in the location and stacking order specified by the corresponding color-coded spheres. The action consists of torques for 3 degrees-of-freedom actuation, and an on-off control input opens and closes the gripper. The environment offers a sparse reward for each block placed at the goal location.

\textbf{Learning levels and option templates:}
We introduce two learning levels with tasks and option templates described in Table~\ref{table:fetch_uber_actions}. 
In this environment, the primitive actions are \emph{not} continuous-space actions. Instead, at the lowest level (level $2$), we expose the agent to options which are implemented with proportional feedback controllers \citep{control_book}. Such simple primitives have the ability to improve exploration and can be easily transferred among different learning scenarios. Further, the hierarchical structure improves the agent's learning speed as compared to a case where the agent has to learn to use the options at the bottom level ($2$) directly (see Appendix \ref{app:fetch} for more details).

% \begin{table}[t]
% \centering
% \begin{tabular}{p{3cm}|p{1.2cm}|p{6cm}}
% \hline
% \footnotesize{Task(s)} & \footnotesize{Learning level} & \footnotesize{Option templates} \\
% \hline \hline
% \footnotesize{Fetch \& Stack $N$ blocks} & \footnotesize{1}  & \footnotesize{\{Place block $i$ at its goal location\}$_{i=1,...,N}$} \\ \hline
% \footnotesize{Place block $i$ at its goal location}   & \footnotesize{2}              & \footnotesize{ \{Reach block $i$, Pick block $i$ \& reach goal,  Release block $i$ \& lift, Do nothing\}$_{i=1,...,N}$} \\ \hline
% \end{tabular}
% \caption{Option templates for fetch and stack.}
% \label{table:fetch_uber_actions}
% \end{table}

% \begin{table}[t]
% \centering
% \begin{tabular}{p{3cm}|p{1.2cm}|p{6cm}}
% \hline
% \footnotesize{Task(s)}       & \footnotesize{Learning level} & \footnotesize{Option templates}           \\ 
% \hline \hline
% \footnotesize{Win game} & \footnotesize{1} & \footnotesize{\{Attack and score goals, Defend\}} \\ \hline
% \footnotesize{Attack and score goals}    & \footnotesize{2}              & \footnotesize{\{Maintain ball possession, Charge to the opponent's goal, Shoot\}}              \\ \hline
% \end{tabular}
% \caption{Option templates for gfootball.}
% \label{table:gfootball_option_templates}
% \end{table}

\begin{table}[t]
\begin{minipage}[t]{.45\linewidth}
\centering
\begin{tabular}{p{2cm}|p{0.5cm}|p{3.1cm}}
% \begin{tabular}{c|c|c}
\hline
\footnotesize{Task(s)} & \footnotesize{level} & \footnotesize{Option templates} \\
\hline \hline
\footnotesize{Fetch \& Stack $N$ blocks} & \footnotesize{1}  & \footnotesize{\{Place block $i$ at its goal location\}$_{i=1,...,N}$} \\ \hline
\footnotesize{Place block $i$ at its goal location}   & \footnotesize{2}              & \footnotesize{ \{Reach block $i$, Pick block $i$ \& reach goal,  Release block $i$ \& lift, Do nothing\}$_{i=1,...,N}$} \\ \hline
\end{tabular}
\caption{Option templates for fetch \& stack.}
\label{table:fetch_uber_actions}
\end{minipage}
\hfill
\begin{minipage}[t]{.45\linewidth}
\centering
\begin{tabular}{p{2cm}|p{0.5cm}|p{2.7cm}}
% \begin{tabular}{c|c|c}
\hline
\footnotesize{Task(s)}       & \footnotesize{level} & \footnotesize{Option templates}           \\ 
\hline \hline
\footnotesize{Win game} & \footnotesize{1} & \footnotesize{\{Attack and score goals, Defend\}} \\ \hline
\footnotesize{Attack and score goals}    & \footnotesize{2}              & \footnotesize{\{Maintain ball possession, Charge to the opponent's goal,  and Shoot\}}              \\ \hline
\end{tabular}
\caption{Option templates for gfootball.}
\label{table:gfootball_option_templates}
\end{minipage}
\vspace{-5mm}
\end{table}

\textbf{Implementation:} We use a standard DQN ~\citep{deep-q-paper} for each level with four hidden layers of 300 neurons each. The inputs to the agent are the 3D coordinates of the different blocks, their goals, and the states of the gripper arm. The agent has 150 and 200 steps for stacking three and four blocks, respectively, in the correct order. We also supply demonstrations to speed up learning.
% \begin{figure}[t!]
% \centering
% \includegraphics[width=.35\textwidth]{Plots/step_block_3_new.png} \qquad
% \includegraphics[width=.35\textwidth]{Plots/step_block_4_new.png}
% \caption{Average reward vs timesteps for solving both learning levels of fetch \& stack. We compare option templates and the baseline, for stacking three and four blocks.%
% }
% \label{fig:plots_fetch}
% \end{figure}

% \begin{figure}[t]
%     \centering
%     \includegraphics[width=0.56\linewidth]{Figures/gfootball_bar_xticks.png}
%     \caption{Comparison of time steps and corresponding average goal difference. Option templates (controlling 11-players), baseline (controlling 11-players), single-player IMPALA~\citep{gfootball}, and single-player DQN~\citep{gfootball} in gfootball.}
%     \label{fig:gfootball_bar_plot}
% \end{figure}

\begin{figure*}[t!]
    \begin{subfigure}[t]{0.29\textwidth}
        \centering
        \includegraphics[width=\linewidth]{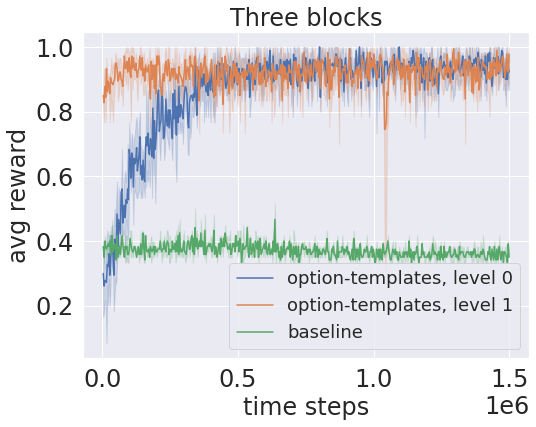}
        \caption{Average reward vs timesteps for option templates and the baseline.}
        \label{fig:plots_fetch}
    \end{subfigure} \hfill 
    \begin{subfigure}[t]{0.29\textwidth}
        \centering
        \includegraphics[width=\linewidth]{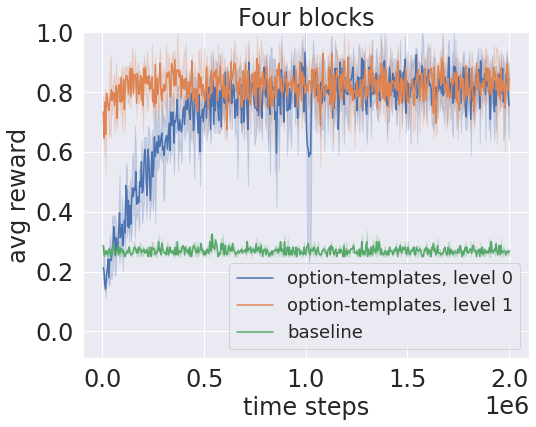}
        \caption{Average reward vs timesteps for option templates and the baseline.}
        \label{fig:plots_fetch_2}
    \end{subfigure} \hfill 
    \begin{subfigure}[t]{0.375\textwidth}
        \centering
        \includegraphics[width=\linewidth]{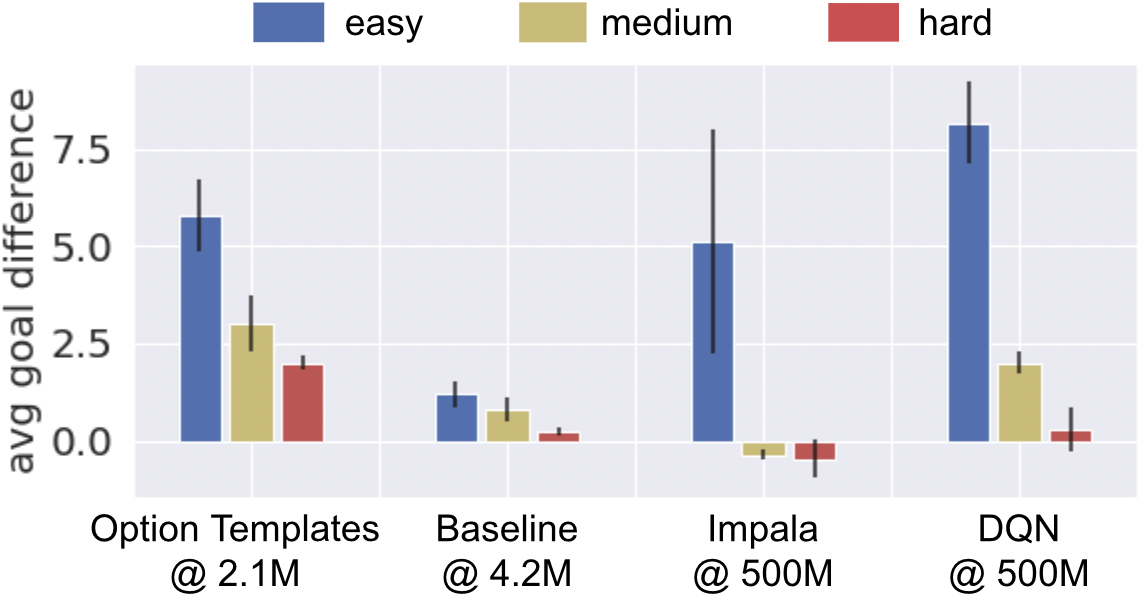}
        \caption{Comparison of time steps and corresponding average goal difference. 
        % Option templates (controlling 11-players), baseline (controlling 11-players), single-player IMPALA~\citep{gfootball}, and single-player DQN~\citep{gfootball} in gfootball.
        }
        \label{fig:gfootball_bar_plot}
    \end{subfigure}
    \caption{Results on the Fetch \& Stack and GFootball environments.}
    \label{fig:env_pictures}
    \vspace{-5mm}
\end{figure*}

\textbf{Baseline:} We consider a baseline that directly exposes the agent to all option templates, at level $2$ of the hierarchy.
Figure \ref{fig:fetch} shows the average rewards as a function of episodes. As can be seen, the baseline only recieves one-third reward since it only succeeds in pushing the bottom block to its location. It does not learn to stack even after twice the number of learning steps as our method. This demonstrates the challenge in learning longer duration tasks without a higher level guiding policy.

\textbf{Comparison with Learning with Demonstrations \citep{demo_RL}: } In learning with demonstrations~\citep{demo_RL}, the authors use demonstration traces, by combining behavioral cloning along with a Q-filter mechanism to speed-up the learning. This allows them to evade an expensive exploration phase in the early stages of the learning.
This method takes upwards of $\mathbf{3.5 \times 10^8}$ and $\underline{8 \times 10^8}$ timesteps to learn stacking of three and four blocks respectively \textbf{which is three orders of magnitude larger than our average learning time} of $\mathbf{4.5 \times 10^5}$ and $\underline{6 \times 10^5}$ timesteps respectively (from 5 random seeds). %
% \vspace{-2mm}

\subsection{Experiments on Multi-Robot Tasks in the GFootball Environment}
\vspace{-1mm}
% \begin{figure}[t]
%     \centering
%     \includegraphics[width=0.56\linewidth]{Figures/gfootball_bar_xticks.png}
%     \caption{Comparison of time steps and corresponding average goal difference. Option templates (controlling 11-players), baseline (controlling 11-players), single-player IMPALA~\citep{gfootball}, and single-player DQN~\citep{gfootball} in gfootball.}
%     \label{fig:gfootball_bar_plot}
% \end{figure}

\textbf{Description of environment: } The Google football (gfootball) environment ~\citep{gfootball} is an 11 vs. 11 game of soccer. The opponents are controlled by an inbuilt game engine with game-play at three levels of difficulty (easy, medium, and hard). An RL agent can control up to 11 players on the left team. The agent provides each player with one of 19 actions such as a direction to move (\textit{e.g.} top, top-right, bottom-left, \textit{etc.}), type of pass (short, long or high), shoot, or toggles for dribbling and sprinting. We provide more details in Appendix \ref{app:gfootball}.
% \todo{There is some description here which needs to go to the Appendix, check the source file}
% The environment returns a unit reward for each goal scored. For our approach, we consider an agent controlling all 11 players; in contrast, the implementations of some baselines we compare to only control one player. Note, this makes the action space much larger, and difficult for exploration. But, makes room for complex strategies to be implemented by the agent.

\textbf{Learning levels and option templates:} We create two learning levels with option templates as given in Table~\ref{table:gfootball_option_templates}. The primitive actions in this environment are \textit{defend}, \textit{maintain ball possession}, \textit{charge} \textit{to the opponent's goal} and \textit{shoot} which are options implemented with simple planers and open-loop controllers. Similarly, we implement option templates at level 1 (see Appendix \ref{app:gfootball}).

\textbf{Implementation: } We train a standard DQN~\citep{deep-q-paper} for each level. The input to every network is a 139 element vector consisting of left and right team states, ball state, score, and one-hot encodings of ball ownership and game mode. For level 1, for all levels of difficulty, we use a network with 2 hidden layers of 500 neurons each. For level 2, we use a network with 5, 6, and 7 hidden layers for easy, medium and hard respectively where each hidden layer has 500 neurons.

\textbf{Baseline: } We consider a baseline where the agent is directly exposed to all the primitive options.
From Figure~\ref{fig:gfootball_bar_plot}, we find that the baseline is unable to compete with option template learning even after twice the number of steps of our method (numerical values of Figure~\ref{fig:gfootball_bar_plot} are in Appendix \ref{app:gfootball}). The error bars represent standard deviation over 5 random seeds.

\textbf{Comparison with IMPALA and DQN from~\citet{gfootball}:} We compare our agent with the agents in~\citet{gfootball}. As can be seen in Figure~\ref{fig:gfootball_bar_plot}, option template learning can achieve similar (in easy) or better (in medium and hard) performance than the IMAPALA and DQN agents in two orders of magnitude fewer steps.

\section{Related Work}
\vspace{-2mm}
% \todo{To be made in sync with Appendix}
% Recent progress in RL has partly been made possible by combining rich function classes like
% deep neural networks with powerful techniques such as Q-learning \citep{main_Q_learning} and actor-critic approaches \citep{main_actor_critic}. In recent years huge strides have been made in training agents for complex tasks, such as for playing Atari games with discrete and continuous action spaces \citep{deep-q-paper,ddpg-paper}. Additionally, RL has achieved human-level performance in games like chess \citep{chess_RL} and Go \citep{GO_RL}. %
% Rest of related work is in the Appendix.

We present a detailed discussion on related work in Appendix \ref{app:related} and summarize closely relevant literature here.

% \textbf{On expert help via primitives or skills}: Various recent works utilize expert help in the form of primitives or skills \citep{ICLR16-hausknecht, ut_austin_robocup, barreto2019option, action_primitives, icra22_primitves, chitnis2020efficient, lynch2020learning}. They do so within the traditional hierarchical framework, \textit{i.e.}, bottom-up (see Figure \ref{fig:sticky-mittens-comp}), where a sub-task is learnt before the policy that uses it. With minimal changes in the implementation of their primitives and skills, the above diverse strategies can benefit from shorter learning times by utilizing our framework to learn top-down.

\textbf{On expert help via primitives or skills}: Various recent works utilize expert help in the form of primitives or skills \citep{ICLR16-hausknecht, ut_austin_robocup, barreto2019option, action_primitives, icra22_primitves, chitnis2020efficient, lynch2020learning}. This takes place via parameterized action spaces \citep{ICLR16-hausknecht}, stitching together independent task schemas (or skills) \citep{ut_austin_robocup, barreto2019option, chitnis2020efficient, lynch2020learning} or learning parameters of action primitves \citep{action_primitives, chitnis2020efficient, icra22_primitves}. In all of these cases, learning takes place within the traditional hierarchical framework, \textit{i.e.}, bottom-up (see Figure \ref{fig:sticky-mittens-comp}), where a sub-task is learnt before the policy that uses it. Our method proposes a framework to learn top-down instead with large-improvements in sample-efficiency over traditional bottom-up learning. With minimal changes in the implementation of their primitives and skills, the above diverse strategies can also benefit with shorter learning times by utilizing our framework to learn top-down.

\textbf{On expert help with humans-in-the-loop}: Other approaches to expert intervention include explicit help with humans-in-the-loop required throughout training \citep{explicit_1, explicit_2}. In contrast, in our method, human effort is only required at the beginning to create the option template hierarchies.

% \textbf{On exploration in high-dimensional tasks}: Orthogonal methods to counter the need for exploration takes place through demonstrations \citep{demo_RL} or \textit{Hindsight Experience Replay} \citep{HER}. By themselves, they are unable to achieve the large degree of reduction in sample-complexity seen with our framework. Yet, alongside expert help, they further improve sample-efficiency as seen in our experiments.

\textbf{On exploration in high-dimensional tasks}: An alternative approach to counter the need for exploration in high dimensional tasks is through the use of demonstrations \citep{demo_RL} for long horizon planning. The intuition is that introducing a degree of \textit{behavioral cloning} of expert demonstrations helps reduce the amount of exploration the agent has to perform. A different approach known as \textit{Hindsight Experience Replay} (HER) \citep{HER} incorporates the goal information into the state, using a failed terminal state as an alternative goal to reward the transitions leading to it. These methods are orthogonal to our approach. By themselves, they are unable to achieve the large degree of reduction in sample-complexity seen with our framework. Yet, alongside expert help, they further improve sample-efficiency as seen in our experiments.

\vspace{-2mm}

\section{Limitations and Conclusion}
\vspace{-2mm}
% We have proposed an approach that incorporates \ops into reinforcement learning. Our experiments show that this strategy can drastically reduce sample complexity compared to existing strategies; importantly, by allowing agents to explore the consequences of complex sub-policies before learning their implementations, it significantly reduces sample complexity compared to a naive strategy that first learns the implementations of all option templates before learning the high-level policy. A key direction for future work is modifying the algorithm to account for potential costs of calling option templates.

We have proposed an approach that incorporates \ops into reinforcement learning. Our experiments show that this strategy can drastically reduce sample complexity by implementing teleportation. This can be a potential limitation. In simulation, implementing teleportation is typically straightforward. Policies may be trained in simulation before being deployed on a real robot. For real-world environments, there are several strategies for implementing teleportation. First, for challenging skills such as grasping an object, teleportation can be implemented via a temporary crutch that simplifies the skill. For example, in the ``sticky-mittens'' experiments (a key motivation for our work); the analog for a grasping robot would be to attach velcro to its grippers and to the objects to make them easy to pick up. Second, teleportation can be implemented via a handcrafted policy that eventually achieves the goal but possibly in a suboptimal way. For instance, teleportation in the Fetch and GFootball environments are implemented using handcrafted policies. While these policies are used to train the RL policy, the RL policy eventually significantly outperforms them. 

\section*{Acknowledgements}
This work was supported in part by ARO W911NF-20-1-0080 and AFRL and DARPA FA8750-18-C-0090. Any opinions, findings, conclusions or recommendations expressed in this material are those of the authors and do not necessarily reflect the views of the Air Force Research Laboratory (AFRL), the Army Research Office (ARO), the Defense Advanced Research Projects Agency (DARPA), the Department of Defense, or the United States Government.

%\newpage
\bibliography{sticky_mittens_references}

\begin{thebibliography}{39}
\providecommand{\natexlab}[1]{#1}
\providecommand{\url}[1]{\texttt{#1}}
\expandafter\ifx\csname urlstyle\endcsname\relax
  \providecommand{\doi}[1]{doi: #1}\else
  \providecommand{\doi}{doi: \begingroup \urlstyle{rm}\Url}\fi

\bibitem[Nair et~al.(2017)Nair, McGrew, Andrychowicz, Zaremba, and
  Abbeel]{demo_RL}
A.~Nair, B.~McGrew, M.~Andrychowicz, W.~Zaremba, and P.~Abbeel.
\newblock Overcoming exploration in reinforcement learning with demonstrations.
\newblock \emph{CoRR}, abs/1709.10089, 2017.
\newblock URL \url{http://arxiv.org/abs/1709.10089}.

\bibitem[Sutton et~al.(1999)Sutton, Precup, and Singh]{options_paper}
R.~S. Sutton, D.~Precup, and S.~Singh.
\newblock Between mdps and semi-mdps: A framework for temporal abstraction in
  reinforcement learning.
\newblock \emph{Artificial Intelligence}, 112\penalty0 (1):\penalty0 181--211,
  1999.
\newblock ISSN 0004-3702.
\newblock \doi{https://doi.org/10.1016/S0004-3702(99)00052-1}.
\newblock URL
  \url{https://www.sciencedirect.com/science/article/pii/S0004370299000521}.

\bibitem[van~den Berg and Gredebäck(2021)]{sticky_mittens_exp}
L.~van~den Berg and G.~Gredebäck.
\newblock The sticky mittens paradigm: A critical appraisal of current results
  and explanations.
\newblock \emph{Developmental Science}, 24\penalty0 (5):\penalty0 e13036,
  September 2021.
\newblock \doi{https://doi.org/10.1111/desc.13036}.
\newblock URL \url{https://onlinelibrary.wiley.com/doi/abs/10.1111/desc.13036}.

\bibitem[Hausknecht and Stone(2016)]{ICLR16-hausknecht}
M.~Hausknecht and P.~Stone.
\newblock Deep reinforcement learning in parameterized action space.
\newblock In \emph{Proceedings of the International Conference on Learning
  Representations (ICLR)}, May 2016.

\bibitem[MacAlpine et~al.(2014)MacAlpine, Depinet, Liang, and
  Stone]{ut_austin_robocup}
P.~MacAlpine, M.~Depinet, J.~Liang, and P.~Stone.
\newblock {UT} austin villa: Robocup 2014 3d simulation league competition and
  technical challenge champions.
\newblock In R.~A.~C. Bianchi, H.~L. Akin, S.~Ramamoorthy, and K.~Sugiura,
  editors, \emph{RoboCup 2014: Robot World Cup {XVIII} [papers from the 18th
  Annual RoboCup International Symposium, Jo{\~{a}}o Pessoa, Brazil, July 15},
  volume 8992 of \emph{Lecture Notes in Computer Science}, pages 33--46.
  Springer, 2014.
\newblock \doi{10.1007/978-3-319-18615-3\_3}.
\newblock URL \url{https://doi.org/10.1007/978-3-319-18615-3\_3}.

\bibitem[Barreto et~al.(2019)Barreto, Borsa, Hou, Comanici, Ayg{\"u}n, Hamel,
  Toyama, Mourad, Silver, Precup, et~al.]{barreto2019option}
A.~Barreto, D.~Borsa, S.~Hou, G.~Comanici, E.~Ayg{\"u}n, P.~Hamel, D.~Toyama,
  S.~Mourad, D.~Silver, D.~Precup, et~al.
\newblock The option keyboard: Combining skills in reinforcement learning.
\newblock \emph{Advances in Neural Information Processing Systems}, 32, 2019.

\bibitem[Dalal et~al.(2021)Dalal, Pathak, and Salakhutdinov]{action_primitives}
M.~Dalal, D.~Pathak, and R.~Salakhutdinov.
\newblock Accelerating robotic reinforcement learning via parameterized action
  primitives.
\newblock \emph{CoRR}, abs/2110.15360, 2021.
\newblock URL \url{https://arxiv.org/abs/2110.15360}.

\bibitem[Nasiriany et~al.(2021)Nasiriany, Liu, and Zhu]{icra22_primitves}
S.~Nasiriany, H.~Liu, and Y.~Zhu.
\newblock Augmenting reinforcement learning with behavior primitives for
  diverse manipulation tasks.
\newblock \emph{arXiv preprint arXiv:2110.03655}, 2021.

\bibitem[Chitnis et~al.(2020)Chitnis, Tulsiani, Gupta, and
  Gupta]{chitnis2020efficient}
R.~Chitnis, S.~Tulsiani, S.~Gupta, and A.~Gupta.
\newblock Efficient bimanual manipulation using learned task schemas.
\newblock In \emph{2020 IEEE International Conference on Robotics and
  Automation (ICRA)}, pages 1149--1155. IEEE, 2020.

\bibitem[Sutton and Barto(2018)]{sutton2018reinforcement}
R.~S. Sutton and A.~G. Barto.
\newblock \emph{Reinforcement learning: An introduction}.
\newblock MIT press, 2018.

\bibitem[Barto and Mahadevan(2003)]{Barto03recentadvances}
A.~G. Barto and S.~Mahadevan.
\newblock Recent advances in hierarchical reinforcement learning.
\newblock \emph{Discrete Event Dynamic Systems Volume 13}, 13:\penalty0 2003,
  2003.

\bibitem[Mnih et~al.(2013)Mnih, Kavukcuoglu, Silver, Graves, Antonoglou,
  Wierstra, and Riedmiller]{deep-q-paper}
V.~Mnih, K.~Kavukcuoglu, D.~Silver, A.~Graves, I.~Antonoglou, D.~Wierstra, and
  M.~Riedmiller.
\newblock Playing atari with deep reinforcement learning, 2013.

\bibitem[Hacohen and Weinshall(2019)]{curriculum_learning}
G.~Hacohen and D.~Weinshall.
\newblock On the power of curriculum learning in training deep networks.
\newblock \emph{CoRR}, abs/1904.03626, 2019.
\newblock URL \url{http://arxiv.org/abs/1904.03626}.

\bibitem[Andreas et~al.(2017)Andreas, Klein, and Levine]{policy_sketches_paper}
J.~Andreas, D.~Klein, and S.~Levine.
\newblock Modular multitask reinforcement learning with policy sketches.
\newblock In \emph{International Conference on Machine Learning}, pages
  166--175. PMLR, 2017.

\bibitem[Behbahani(2018)]{craft_env_github_visualization_repo}
F.~Behbahani.
\newblock Craft environment.
\newblock \url{https://github.com/Feryal/craft-env}, 2018.

\bibitem[Konda and Tsitsiklis(2001)]{main_actor_critic}
V.~Konda and J.~Tsitsiklis.
\newblock Actor-critic algorithms.
\newblock \emph{Society for Industrial and Applied Mathematics}, 42, 04 2001.

\bibitem[Lanier(2019)]{gym_fetch_stack}
J.~B. Lanier.
\newblock \emph{Curiosity-driven multi-criteria hindsight experience replay}.
\newblock University of California, Irvine, 2019.

\bibitem[Astrom and Murray(2008)]{control_book}
K.~J. Astrom and R.~M. Murray.
\newblock \emph{Feedback Systems: An Introduction for Scientists and
  Engineers}.
\newblock Princeton University Press, USA, 2008.
\newblock ISBN 0691135762.

\bibitem[Kurach et~al.(2019)Kurach, Raichuk, Sta{\'n}czyk, Zajac, Bachem,
  Espeholt, Riquelme, Vincent, Michalski, Bousquet, et~al.]{gfootball}
K.~Kurach, A.~Raichuk, P.~Sta{\'n}czyk, M.~Zajac, O.~Bachem, L.~Espeholt,
  C.~Riquelme, D.~Vincent, M.~Michalski, O.~Bousquet, et~al.
\newblock Google research football: A novel reinforcement learning environment.
\newblock \emph{arXiv preprint arXiv:1907.11180}, 2019.

\bibitem[Lynch et~al.(2020)Lynch, Khansari, Xiao, Kumar, Tompson, Levine, and
  Sermanet]{lynch2020learning}
C.~Lynch, M.~Khansari, T.~Xiao, V.~Kumar, J.~Tompson, S.~Levine, and
  P.~Sermanet.
\newblock Learning latent plans from play.
\newblock In \emph{Conference on robot learning}, pages 1113--1132. PMLR, 2020.

\bibitem[Li et~al.(2022)Li, Peng, and Zhou]{explicit_1}
Q.~Li, Z.~Peng, and B.~Zhou.
\newblock Efficient learning of safe driving policy via human-ai copilot
  optimization.
\newblock \emph{arXiv preprint arXiv:2202.10341}, 2022.

\bibitem[Spencer et~al.(2022)Spencer, Choudhury, Barnes, Schmittle, Chiang,
  Ramadge, and Srinivasa]{explicit_2}
J.~Spencer, S.~Choudhury, M.~Barnes, M.~Schmittle, M.~Chiang, P.~Ramadge, and
  S.~Srinivasa.
\newblock Expert intervention learning.
\newblock \emph{Autonomous Robots}, 46\penalty0 (1):\penalty0 99--113, 2022.

\bibitem[Andrychowicz et~al.(2017)Andrychowicz, Wolski, Ray, Schneider, Fong,
  Welinder, McGrew, Tobin, Abbeel, and Zaremba]{HER}
M.~Andrychowicz, F.~Wolski, A.~Ray, J.~Schneider, R.~Fong, P.~Welinder,
  B.~McGrew, J.~Tobin, P.~Abbeel, and W.~Zaremba.
\newblock Hindsight experience replay.
\newblock \emph{CoRR}, abs/1707.01495, 2017.
\newblock URL \url{http://arxiv.org/abs/1707.01495}.

\bibitem[Watkins and Dayan(1992)]{main_Q_learning}
C.~Watkins and P.~Dayan.
\newblock Technical note: Q-learning.
\newblock \emph{Machine Learning}, 8:\penalty0 279--292, 05 1992.
\newblock \doi{10.1007/BF00992698}.

\bibitem[Akhmetzyanov et~al.(2020)Akhmetzyanov, Yagfarov, Gafurov, Ostanin, and
  Klimchik]{ddpg-paper}
A.~Akhmetzyanov, R.~Yagfarov, S.~Gafurov, M.~Ostanin, and A.~Klimchik.
\newblock Continuous control in deep reinforcement learning with direct policy
  derivation from q network.
\newblock In T.~Ahram, R.~Taiar, V.~Gremeaux-Bader, and K.~Aminian, editors,
  \emph{Human Interaction, Emerging Technologies and Future Applications II},
  pages 168--174, Cham, 2020. Springer International Publishing.

\bibitem[Hsu(2002)]{chess_RL}
F.-H. Hsu.
\newblock \emph{Behind Deep Blue: Building the Computer That Defeated the World
  Chess Champion}.
\newblock Princeton University Press, USA, 2002.
\newblock ISBN 0691090653.

\bibitem[Silver et~al.(2016)Silver, Huang, Maddison, Guez, Sifre, Driessche,
  Schrittwieser, Antonoglou, Panneershelvam, Lanctot, Dieleman, Grewe, Nham,
  Kalchbrenner, Sutskever, Lillicrap, Leach, Kavukcuoglu, Graepel, and
  Hassabis]{GO_RL}
D.~Silver, A.~Huang, C.~Maddison, A.~Guez, L.~Sifre, G.~Driessche,
  J.~Schrittwieser, I.~Antonoglou, V.~Panneershelvam, M.~Lanctot, S.~Dieleman,
  D.~Grewe, J.~Nham, N.~Kalchbrenner, I.~Sutskever, T.~Lillicrap, M.~Leach,
  K.~Kavukcuoglu, T.~Graepel, and D.~Hassabis.
\newblock Mastering the game of go with deep neural networks and tree search.
\newblock \emph{Nature}, 529:\penalty0 484--489, 01 2016.
\newblock \doi{10.1038/nature16961}.

\bibitem[Levy et~al.(2017)Levy, Jr., and Saenko]{HRL_actor_critic}
A.~Levy, R.~P. Jr., and K.~Saenko.
\newblock Hierarchical actor-critic.
\newblock \emph{CoRR}, abs/1712.00948, 2017.
\newblock URL \url{http://arxiv.org/abs/1712.00948}.

\bibitem[Yang et~al.(2020)Yang, Yuan, Zhu, Yu, and Li]{HRL_robotics}
C.~Yang, K.~Yuan, Q.~Zhu, W.~Yu, and Z.~Li.
\newblock Multi-expert learning of adaptive legged locomotion.
\newblock \emph{CoRR}, abs/2012.05810, 2020.
\newblock URL \url{https://arxiv.org/abs/2012.05810}.

\bibitem[Marzari et~al.(2021)Marzari, Pore, Dall'Alba, Aragon{-}Camarasa,
  Farinelli, and Fiorini]{HRL_fetch_place}
L.~Marzari, A.~Pore, D.~Dall'Alba, G.~Aragon{-}Camarasa, A.~Farinelli, and
  P.~Fiorini.
\newblock Towards hierarchical task decomposition using deep reinforcement
  learning for pick and place subtasks.
\newblock \emph{CoRR}, abs/2102.04022, 2021.
\newblock URL \url{https://arxiv.org/abs/2102.04022}.

\bibitem[Winder et~al.(2020)Winder, Milani, Landen, Oh, Parr, Squire,
  desJardins, and Matuszek]{PALM}
J.~Winder, S.~Milani, M.~Landen, E.~Oh, S.~Parr, S.~Squire, M.~desJardins, and
  C.~Matuszek.
\newblock Planning with abstract learned models while learning transferable
  subtasks.
\newblock \emph{Proceedings of the AAAI Conference on Artificial Intelligence},
  34\penalty0 (06):\penalty0 9992--10000, Apr. 2020.
\newblock \doi{10.1609/aaai.v34i06.6555}.
\newblock URL \url{https://ojs.aaai.org/index.php/AAAI/article/view/6555}.

\bibitem[Jothimurugan et~al.(2020)Jothimurugan, Bastani, and
  Alur]{abstract_val_iteration}
K.~Jothimurugan, O.~Bastani, and R.~Alur.
\newblock Abstract value iteration for hierarchical reinforcement learning.
\newblock \emph{CoRR}, abs/2010.15638, 2020.
\newblock URL \url{https://arxiv.org/abs/2010.15638}.

\bibitem[Abel et~al.(2020)Abel, Umbanhowar, Khetarpal, Arumugam, Precup, and
  Littman]{pmlr-v108-abel20a}
D.~Abel, N.~Umbanhowar, K.~Khetarpal, D.~Arumugam, D.~Precup, and M.~Littman.
\newblock Value preserving state-action abstractions.
\newblock In S.~Chiappa and R.~Calandra, editors, \emph{Proceedings of the
  Twenty Third International Conference on Artificial Intelligence and
  Statistics}, volume 108 of \emph{Proceedings of Machine Learning Research},
  pages 1639--1650. PMLR, 26--28 Aug 2020.
\newblock URL \url{https://proceedings.mlr.press/v108/abel20a.html}.

\bibitem[Yang et~al.(2021)Yang, Inala, Bastani, Pu, Solar-Lezama, and
  Rinard]{prog_sys}
Y.~Yang, J.~Inala, O.~Bastani, Y.~Pu, A.~Solar-Lezama, and M.~Rinard.
\newblock Program synthesis guided reinforcement learning, 02 2021.

\bibitem[Gopalan et~al.(2017)Gopalan, desJardins, Littman, MacGlashan, Squire,
  Tellex, Winder, and Wong]{Gopalan2017PlanningWA}
N.~Gopalan, M.~desJardins, M.~L. Littman, J.~MacGlashan, S.~Squire, S.~Tellex,
  J.~Winder, and L.~L.~S. Wong.
\newblock Planning with abstract markov decision processes.
\newblock In \emph{ICAPS}, 2017.

\bibitem[Jothimurugan et~al.(2021)Jothimurugan, Bansal, Bastani, and
  Alur]{DBLP:journals/corr/abs-2106-13906}
K.~Jothimurugan, S.~Bansal, O.~Bastani, and R.~Alur.
\newblock Compositional reinforcement learning from logical specifications.
\newblock \emph{CoRR}, abs/2106.13906, 2021.
\newblock URL \url{https://arxiv.org/abs/2106.13906}.

\bibitem[Garrett et~al.(2021)Garrett, Chitnis, Holladay, Kim, Silver,
  Kaelbling, and Lozano-P{\'e}rez]{tamp}
C.~R. Garrett, R.~Chitnis, R.~Holladay, B.~Kim, T.~Silver, L.~P. Kaelbling, and
  T.~Lozano-P{\'e}rez.
\newblock Integrated task and motion planning.
\newblock \emph{Annual review of control, robotics, and autonomous systems},
  4:\penalty0 265--293, 2021.

\bibitem[Levy et~al.(2017)Levy, Konidaris, Platt, and Saenko]{new_comparison}
A.~Levy, G.~Konidaris, R.~Platt, and K.~Saenko.
\newblock Learning multi-level hierarchies with hindsight.
\newblock \emph{arXiv preprint arXiv:1712.00948}, 2017.

\bibitem[Yang et~al.(2021)Yang, Ji, Wu, Lai, Wei, Liu, and
  Setchi]{new_comparison_2}
X.~Yang, Z.~Ji, J.~Wu, Y.-K. Lai, C.~Wei, G.~Liu, and R.~Setchi.
\newblock Hierarchical reinforcement learning with universal policies for
  multistep robotic manipulation.
\newblock \emph{IEEE Transactions on Neural Networks and Learning Systems},
  2021.

\end{thebibliography}
% \bibliographystyle{corlabbrvnat}

%\newpage
\appendix
\newpage
\section{Appendix}
\subsection{Related Work} 
\label{app:related}
Recent progress in RL has partly been made possible by combining rich function classes like
deep neural networks with powerful techniques such as Q-learning \citep{main_Q_learning} and actor-critic approaches \citep{main_actor_critic}. 
In recent years huge strides have been made in training agents for complex tasks, such as for playing Atari games with discrete and continuous action spaces \citep{deep-q-paper,ddpg-paper}. Additionally, RL has achieved human-level performance in games like chess \citep{chess_RL} and Go \citep{GO_RL}. %

However, it is becoming increasingly clear that simple exploration is not enough to circumvent the curse of dimensionality in environments with long horizons and sparse rewards.  
As a remedy the broad area of Hierarchical Reinforcement Learning (HRL) attempts to decompose RL problems into multiple levels of abstraction--- temporal, spatial, or otherwise.
Many works deploy separate policies over different time horizons and action spaces \citep{Barto03recentadvances, HRL_actor_critic, HRL_robotics, HRL_fetch_place}. 
Temporal abstraction in planning can be traced back at least to \citet{options_paper}, where the options were introduced to refer to lower level policies. In most existing research in hierarchical RL, learning a sub-task precedes the learning of a policy which uses it. This includes \citep{PALM, abstract_val_iteration,pmlr-v108-abel20a, prog_sys, Gopalan2017PlanningWA}. This paper is set apart by the fact it does not rely on an apriori knowledge of the starting distributions of the sub-tasks as required by some related literature on hierarchical RL, such as \citep{DBLP:journals/corr/abs-2106-13906}. We further elaborate on literature relevant to the two important facets of our approach below.

\textbf{On expert help via primitives or skills}: Various recent works utilize expert help in the form of primitives or skills \citep{ICLR16-hausknecht, ut_austin_robocup, barreto2019option, action_primitives, icra22_primitves, chitnis2020efficient, lynch2020learning}. This takes place via parameterized action spaces \citep{ICLR16-hausknecht}, stitching together independent task schemas (or skills) \citep{ut_austin_robocup, barreto2019option, chitnis2020efficient, lynch2020learning} or learning parameters of action primitves \citep{action_primitives, chitnis2020efficient, icra22_primitves}. In all of these cases, learning takes place within the traditional hierarchical framework, \textit{i.e.}, bottom-up (see Figure \ref{fig:sticky-mittens-comp}), where a sub-task is learnt before the policy that uses it. Our method proposes a framework to learn top-down instead with large-improvements in sample-efficiency over traditional bottom-up learning. With minimal changes in the implementation of their primitives and skills, the above diverse strategies can also benefit with shorter learning times by utilizing our framework to learn top-down.

\textbf{On expert help with humans-in-the-loop}: Other approaches to expert intervention include explicit help with humans-in-the-loop required throughout training \citep{explicit_1, explicit_2}. In contrast, in our method, human effort is only required at the beginning to create the option template hierarchies. 

\textbf{On hierarchical learning in robotics}: Task and motion planning (TAMP) methods \citep{tamp} are alternate hierarchical approaches for tackling long-horizon problems in robotics. They generally highlight the interplay of motion-level and task-level planning, with the task-level planner constraining the motion-level planner. In contrast, our approach allows the task-level planner to learn using feasible paths which can be satisfied by the motion-level planner. This allows us to achieve high reward with few samples.

\textbf{On exploration in high-dimensional tasks}: An alternative approach to counter the need for exploration in high dimensional tasks is through the use of demonstrations \citep{demo_RL} for long horizon planning. The intuition is that introducing a degree of \textit{behavioral cloning} of expert demonstrations helps reduce the amount of exploration the agent has to perform. A different approach known as \textit{Hindsight Experience Replay} (HER) \citep{HER} incorporates the goal information into the state, using a failed terminal state as an alternative goal to reward the transitions leading to it. These methods are orthogonal to our approach. By themselves, they are unable to achieve the large degree of reduction in sample-complexity seen with our framework. Yet, alongside expert help, they further improve sample-efficiency as seen in our experiments.

\subsection{Additional Details of Experiments on the Craft Environment}\label{app:craft}

\subsubsection{Results on the \textit{get gold} Task.}
\begin{table}[h]
\footnotesize
\begin{minipage}{.35\linewidth}
    \centering
    \begin{tabular}{c|c}
    \hline
    Task        & Policy Sketch                                                 \\ 
    \hline \hline 
    get wood    & -                                                             \\
    get iron    & -                                                             \\ \hline
    make bridge & get wood $\rightarrow$ get iron                             \\ 
                & $\rightarrow$ use factory                                  \\
    \hline     
    get gold     & make bridge $\rightarrow$ \\
    & use bridge on water    \\ 
    \hline
    \end{tabular}
\end{minipage}
\begin{minipage}{.6\linewidth}
    \centering
    \begin{tabular}{c|c|l}
    \hline
    Task       & Learning & Option templates          \\ 
                & level         &                       \\
    \hline \hline
    get gold    & 1              & \{give bridge, primitive actions\}              \\ \hline
    make bridge & 2              & \{give wood, give iron, primitive  \\ 
                &               &                             actions\} \\ \hline
    get wood   & 3              & \{primitive actions\}                     \\ \hline
    get iron   & 3              & \{primitive actions\}                    \\ \hline
    \end{tabular}
\end{minipage}
\label{table:craft_tasks_and_uber_actions_for_gold}
\caption{Get gem hierarchical task: \textbf{[LEFT]} policy sketches
\citep{policy_sketches_paper} \& \textbf{[RIGHT]} option templates for each task in the hierarchy. The order of the rows represent the learning order in the two alternatives.} 
\end{table}

\begin{table}[h]
\footnotesize
\centering
\begin{tabular}{c|c|c}
Task         & \multicolumn{2}{c}{Episodes}                                                              \\ 
\cline{2-3} & Curriculum learning & Option templates \\
 & ~\citep{policy_sketches_paper} & (Ours) \\ 
\hline \hline 
get gold & $>2.65 \times 10^6$  & 10516.0 \scriptsize{$\pm$ 2833.0}\\
make bridge & $>1.8 \times 10^6$  & 7974.0 \scriptsize{$\pm$ 1853.0}\\
\end{tabular}
\caption{Comparison of total episodes (and standard deviations over ten random seeds) to train an agent to solve the \textit{get gem} task via option templates and curriculum learning~\citep{policy_sketches_paper}.}
\label{table:craft_results_gold}
\end{table}

\begin{figure}[H]
\centering
\includegraphics[width=.32\textwidth]{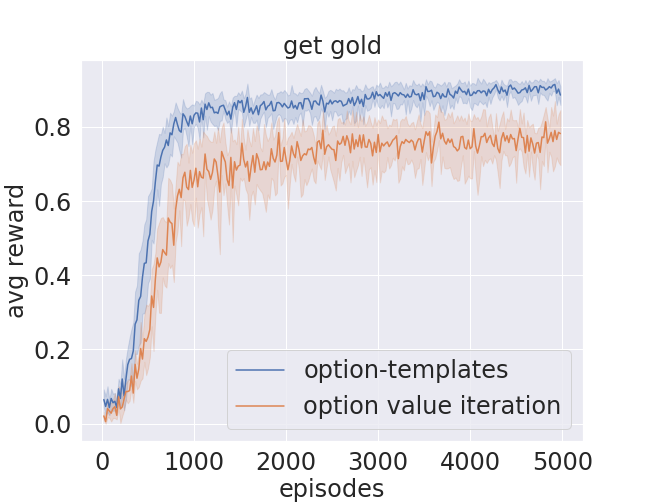}
\includegraphics[width=.32\textwidth]{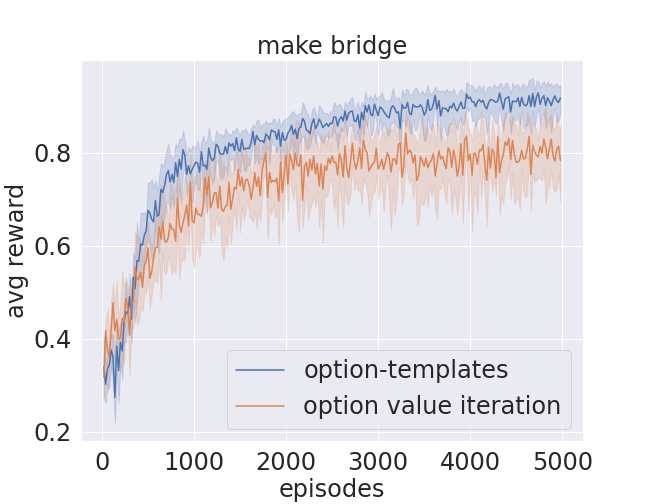}
\caption{Average reward vs episodes for solving each hierarchical sub-task. We compare option templates (our method) and option-value iteration (baseline) for the task \textit{get gold} in craft environment.} 
\label{fig:plots_craft_gold}
\end{figure}

\subsubsection{Additional Environment Details}
For a fair comparison with~\citep{policy_sketches_paper}, we set the maximum steps in the environment for \textit{get wood} and \textit{get iron} to 100; for \textit{make stick} to 200; for \textit{make bridge} to 300; for \textit{make axe} and \textit{get gold} to 400; and for \textit{get gem} to 500.

\subsubsection{Hyperparameters}
The Adam optimizer was used to train all networks. The batch size was always set to the size of the memory.  The exploration ($\epsilon$) is exponentially reduced.
For all tasks, we use a discount factor ($\gamma$) of $0.99$ for primitive actions and $0.99^{d}$ for option templates. The exponent ($d$) is the expected task horizon and is reported in Table~\ref{table:hyper_craft}. Other hyperparameters are also given in Table~\ref{table:hyper_craft}.

In option value iteration, we use the exact task horizon which is available to the agent since options are learnt bottom-up. The other hyperparameters in option value iteration are the same as those reported for option templates.

\begin{table}[h]
    \centering
    \begin{tabular}{c|c|c|c|c|c}
        Parameters & \textit{get gem} & \textit{make axe} & \textit{make stick} & \textit{get gold} & \textit{make bridge}\\ \hline \hline
        expected task horizon ($d$) & 100 & 50 & 40 & 100 & 50 \\ \hline
        memory size (transitions) & \multicolumn{5}{c}{20$\times d$} \\ \hline
        learning frequency (steps) & \multicolumn{5}{c}{20$\times d$} \\ \hline
        learning rate & \multicolumn{5}{c}{0.001} \\ \hline
    \end{tabular}
    \caption{Hyperparameters for option templates in craft environment.}
    \label{table:hyper_craft}
\end{table}

\subsubsection{Discussion}
Figures \ref{fig:plots_craft} and \ref{fig:plots_craft_gold} show that option templates obtain a higher average reward than option-value iteration at all levels. In particular, at level 2, the two curves appear closer since the option templates curve converges to a lower value than in levels 1 and 3, likely due to the tasks on level 2 being
more difficult. We believe the gap occurs because policies at the levels below the one currently being learned typically do not perform perfectly in option-value iteration. Further, these errors may “confuse” training the option in option-value iteration. Whereas in our approach which uses option templates which do not make these errors, we avoid the aforementioned issue.

\subsection{Additional Details of Experiments on the Fetch and Stack Environment}\label{app:fetch}
The Fetch and Stack environment is implemented inside the MuJoCo physics engine designed to simulate multi-body interactions including contacts, joints and collision. Unlike craft, this environment does not permit the implementation of option-templates as part of its action space. We circumvent this problem by adding action primitives using simple state-feedback controllers. Each option-template is implemented in this fashion, for different learning levels (see Table \ref{table:fetch_uber_actions}), until the lowest level is reached. The agent treats the state feedback controllers as option-templates during training phase. As intended, the agent only gets access to read the states after the option-templates reach the termination condition. The option templates can also terminate after a timeout (80 steps for level 1 and 30 steps for level 2).

\subsubsection{Hyperparameters}
The Adam optimizer was used to train all networks. The batch size was always set to $20$\% of the memory size. The exploration ($\epsilon$) is exponentially reduced. %
All other hyperparameters are given in Table~\ref{table:hyper_fetch}. The baseline used the same hyperparameters as option templates.
\begin{table}[h]
    \centering
    \begin{tabular}{c|c|c}
        Parameters & level 1 & level 2\\ \hline \hline 
        discount factor ($\gamma$) & 0.2 & 0.8\\ \hline
        memory size (transitions) & \multicolumn{2}{c}{100}\\ \hline
        learning frequency & \multicolumn{2}{c}{every 1500 steps for 3 blocks} \\
         & \multicolumn{2}{c}{every 2000 steps for 4 blocks}\\ \hline
        learning rate & \multicolumn{2}{c}{0.001}\\ \hline
    \end{tabular}
    \caption{Hyperparameters for option templates in fetch and stack environment.}
    \label{table:hyper_fetch}
\end{table}

\subsubsection{Comparison to Learning multi-level hierarchies with hindsight \citep{new_comparison}}
We also compared our method with \citet{new_comparison}’s Hindsight Actor Critic (HAC) algorithm in the Fetch and Stack environment on the three block stacking task and clearly notice that it cannot learn to stack even two blocks in more than 10 times our learning time.

HAC \citep{new_comparison} obtain an average reward of only $0.138$ after $(67.2 \pm 0.1) \times 10^5$  steps (obtained over 5 random seeds). In comparison, we obtain an average reward of 1 after a learning duration of only $(4.5 \pm 0.1) \times 10^5$ timesteps.

\begin{figure}[H]
\centering
\includegraphics[width=.55\textwidth]{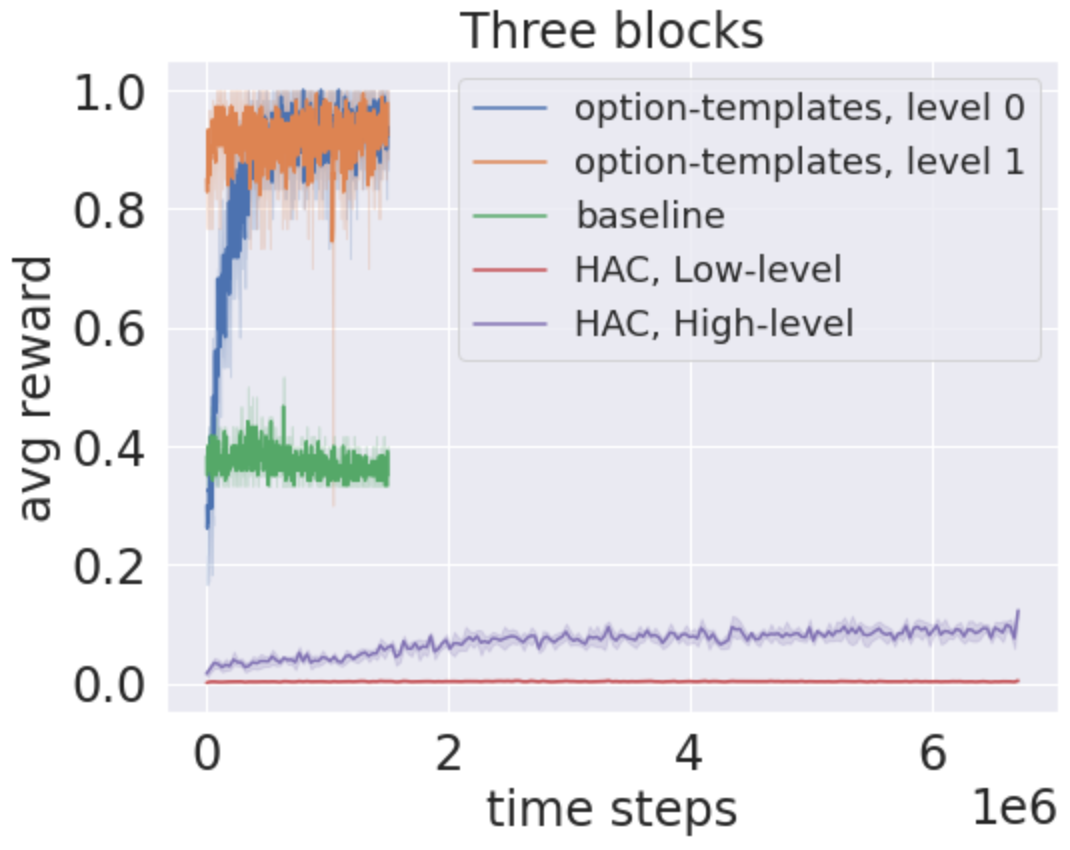}
\caption{Average reward vs episodes for all levels with option templates (our method), the baseline and HAC \citep{new_comparison} in the Fetch and Stack environment.} 
\label{fig:new_comparison}
\end{figure}

We note that, we use the high-level / low-level learning strategy described in \citet{new_comparison_2} and only mention the high-level learning time (which is much lower than the low-level learning time) for \citet{new_comparison} here for a fair comparison. We plot the variation of rewards at all levels with our method, baseline and HAC \citep{new_comparison} in Figure \ref{fig:new_comparison}.

Further, the method proposed in \citet{new_comparison_2} called Universal Option Framework is shown to outperform \citet{new_comparison}’s Hindsight Actor Critic. Even with this boost, according to \citet{new_comparison_2}, their method attains a 0.7 average reward at the high-level only after $(480 \pm 3.2) \times 10^5$ timesteps.

\subsection{Additional Details of Experiments on the GFootball Environment}\label{app:gfootball}

The GFootball environment returns a unit reward for each goal scored. For our approach, we consider an agent controlling all 11 players; in contrast, the implementations of some baselines we compare to only control one player. The task where the agent must control all 11 players is significantly more challenging, and we are unaware of any 11-player baseline. This is because, our action space is much larger, and difficult for exploration. 
% But, we note that this makes room for complex strategies to be implemented by our agent. 

Similar to fetch and stack, since the environment does not have any teleportation support, we implement option templates using simple planners and open-loop controllers (described below). We also provide the numerical values plotted in Figure 6 in Table~\ref{table:gfootball_numbers}.

\begin{table}[t]
\footnotesize
\centering
\begin{tabular}{c|c|c|c|c|c|c}
Method & \multicolumn{2}{c|}{Easy} & \multicolumn{2}{c|}{Medium} & \multicolumn{2}{c}{Hard} \\
\cline{2-7} & avg. goal & steps  & avg. goal & steps & avg. goal & steps  \\ 
 & difference & & difference & & difference & \\
\hline \hline
Option   & $5.79 \pm 0.92$    & $0.3$ M    & $3.0 \pm 0.71$      & $0.3$ M      & $2.0 \pm 0.18$    & $0.3$ M    \\ 
Templates & & (win game) & & (win game)      & & (win game)     \\ 
 & & + $1.8$ M & & + $1.8$ M & & + $1.8$ M \\
 & & (attack) & & (attack)      & & (attack)     \\ 
 & & = $2.1$ M & & = $2.1$ M & & = $2.1$ M \\ 
\hline 
Baseline          & $1.2 \pm 0.22$    & $4.2$ M     & $0.8 \pm 0.33$      & $4.2$ M      & $0.26 \pm 0.1$    & $4.2$ M     \\ \hline
1-Player  & $5.14 \pm 2.88$ & $500$ M  & $-0.36 \pm 0.11$      & $500$ M   & $-0.47 \pm 0.48$    & $500$ M  \\ 
Impala ~\citep{gfootball} & & & & & & \\ \hline
1-Player DQN & $8.16 \pm 1.05$    & $500$ M  & $2.01 \pm 0.27$      & $500$ M   & $0.27 \pm 0.56$    & $500$ M  \\
DQN~\citep{gfootball} & & & & & &
\end{tabular}
\caption{Numerical values of Figure~\ref{fig:gfootball_bar_plot}: comparison of time steps and corresponding average goal difference (including standard deviation across 5 random seeds) for option templates (controlling 11-players), baseline (controlling 11-players), and single-player agents~\citep{gfootball}.} %
\label{table:gfootball_numbers}
\end{table}

\subsubsection{Additional Environment and Option Template Details}
The dimensions of the football field is bounded by the following limits :  $[-1,1]$ along the x-coordinate, and $[-0.42, 0.42]$ along the y-coordinate. 
Of the 19 available actions, there are 8 movement actions, one for each of the following directions - \{top, top-right, right, bottom-right, bottom, bottom-left, left, and top-left \}. There are three actions for passing - \{ long pass, short pass and high pass\}, and one for shooting. There are actions to toggle the following modes - \{ sprinting, dribbling and movement\}. The environment permits an action to let the game engine pick a default move.

The implementations of the option templates in Table~\ref{table:gfootball_option_templates} are described below :

\textbf{Charge to the opponent's goal: }This option template makes the ball controlling player on the agent's team run in a straight line to the center of the opponent's goal. The precondition for this option template being that the agent's team has the ball. The option template terminates when a player with the ball is within a distance of 0.3 units from the goal (1,0). If the precondition is satisfied, the player sprints in one of the above 8 directions, depending on the angle between the ball controlling player and the opponent's goal. 

\textbf{Maintain ball possession: }This option template enables the agent's team to maintain ball possession by either passing the ball to a teammate or move slowly towards the goal while avoiding opponent players who can intercept the ball. When no opponent interceptors are identified ahead of the ball controlling player, in one of 5 relevant directions (top, top-right, right, bottom-right or bottom), the player moves in the identified free direction. Which by default, is towards the opponent's side of the field. This attacker is supported by two players (wings). Otherwise, if no free space is identified, the ball controlling player passes the ball to a teammate with the least number of opponents around him who can intercept the pass. The choice of action (short, long, and high - pass) is based on the  distance between the two players.

\textbf{Shoot: }This option template is the same as the environment action to shoot a goal.

\textbf{Attack and score goals: }This option template is a composition of \textit{charge to the opponent's goal}, \textit{maintain ball possession} and \textit{shoot}. When the x coordinate of the ball controlling player is greater than or equal to 0.5, the ball controlling player charges to the goal. When this attacking player is within 0.25 units from the goal, the player takes a shot. Otherwise, the agent's team simply maintains ball possession.

\textbf{Defend: }This option template replicates the inbuilt game engine to block goal attempts by the other team and get ball possession.

All of the above option templates will also terminate after 200 steps. Additionally, for level 1, we mask all but the one hot encoding of ball ownership in the input. Further, while it is possible to combine \textit{defend} with \textit{attack and score goals} to play the game, they are unable to achieve similar performance levels as compared to the baseline or option templates.

\subsubsection{Hyperparameters}
The Adam optimizer was used to train all networks. The batch size was always set to $40$\% of the memory size. The exploration ($\epsilon$) is exponentially reduced.
All other hyperparameters are given in Table~\ref{table:hyper_gfootball_OT}. The hyperparameters for level 2 in Table~\ref{table:hyper_gfootball_OT} were also utilized in the baseline.

\begin{table}[h]
    \centering
    \begin{tabular}{c|c|c|c|c|c|c}
        Parameters &  \multicolumn{3}{c|}{level 1} & \multicolumn{3}{c}{level 2}\\
        \cline{2-7} & easy & medium & hard & easy & medium & hard\\ \hline \hline 
        discount factor ($\gamma$) & \multicolumn{3}{c|}{0.9} & 0.93 & 0.93 & 0.96 \\ \hline
        memory size (transitions)  & \multicolumn{3}{c|}{6000} & 6000 & 3000 & 3000\\ \hline
        learning frequency  & \multicolumn{3}{c|}{every 3000 steps} & every 3000 steps & \multicolumn{2}{c}{every 30,000 steps} \\ \hline 
        learning rate  & \multicolumn{3}{c|}{0.001} & 0.001 & 0.01 & 0.005\\ \hline
    \end{tabular}
    \caption{Hyperparameters for option templates in gfootball environment.}
    \label{table:hyper_gfootball_OT}
\end{table}

\end{document}